\documentclass[pdflatex,sn-mathphys-num]{sn-jnl}

\usepackage{graphicx}%
\usepackage{multirow}%
\usepackage{amsmath,amssymb,amsfonts}%
\usepackage{amsthm}%
\usepackage{mathrsfs}%
\usepackage[title]{appendix}%
\usepackage{xcolor}%
\usepackage{textcomp}%
\usepackage{manyfoot}%
\usepackage{booktabs}%
\usepackage{algorithm}%
\usepackage{algorithmicx}%
\usepackage{algpseudocode}%
\usepackage{listings}%

\usepackage[strings]{underscore}
\usepackage{ulem}
\usepackage{adjustbox}
\usepackage{bm}
\usepackage{mathtools}
\usepackage{enumitem}
\usepackage{multirow}
\usepackage{makecell}
\usepackage{threeparttable}
\usepackage{subcaption}
\usepackage[figuresright]{rotating}
\usepackage[listings, skins, breakable, theorems]{tcolorbox}

\newtcolorbox[use counter=textcounter]{tcbtextbox}[3][]{
    title={Text \thetcbcounter:\quad#2},
    halign title=flush center,
    enhanced,
    #3,
    #1
}

\newcommand{\ul}[1]{\underline{#1}}

\theoremstyle{thmstyleone}%
\newtheorem{theorem}{Theorem}
\newtheorem{proposition}[theorem]{Proposition}%

\theoremstyle{thmstyletwo}%

\theoremstyle{thmstylethree}%

\begin{document}

\title{Toward Formalizing LLM-Based Agent Designs through Structural Context Modeling and Semantic Dynamics Analysis}


\author*[1]{\fnm{Haoyu} \sur{Jia}}\email{jia@jsk.t.u-tokyo.ac.jp}

\author[1]{\fnm{Kento} \sur{Kawaharazuka}}

\author[1]{\fnm{Kei} \sur{Okada}}

\affil[1]{\orgdiv{Graduate School of Information Science and Technology}, \orgname{The University of Tokyo}, \orgaddress{\country{Japan}}}


\abstract{
Current research on large language model (LLM) agents is fragmented: discussions of conceptual frameworks and methodological principles are frequently intertwined with low-level implementation details, causing both readers and authors to lose track amid a proliferation of superficially distinct concepts.
We argue that this fragmentation largely stems from the absence of an analyzable, self-consistent formal model that enables implementation-independent characterization and comparison of LLM agents.
To address this gap, we propose the \texttt{Structural Context Model}, a formal model for analyzing and comparing LLM agents from the perspective of context structure.
Building upon this foundation, we introduce two complementary components that together span the full lifecycle of LLM agent research and development: (1) a declarative implementation framework; and (2) a sustainable agent engineering workflow, \texttt{Semantic Dynamics Analysis}.
The proposed workflow provides principled insights into agent mechanisms and supports rapid, systematic design iteration. We demonstrate the effectiveness of the complete framework on dynamic variants of the monkey-banana problem, where agents engineered using our approach achieve up to a 32 percentage points improvement in success rate on the most challenging setting.
}




\maketitle


\section{Introduction}

\begin{figure}[ht!]
\centering
\includegraphics[width=\textwidth]{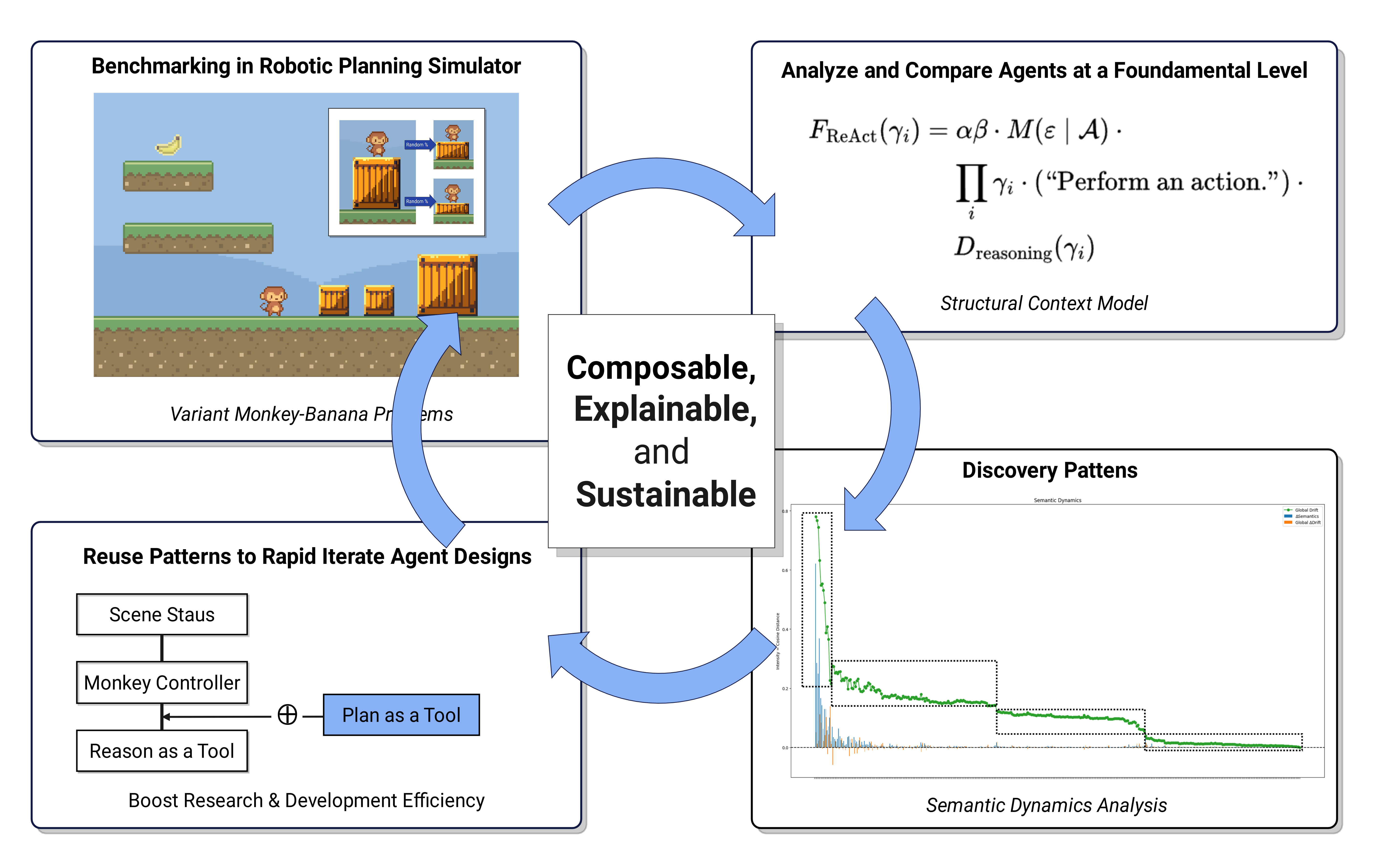}
\caption{
We present a composable, explainable, and sustainable framework for LLM agent research and development.
Starting from the \texttt{Structural Context Model}, agents are analyzed and compared at a fundamental level, enabling \texttt{Semantic Dynamics Analysis} to discover reusable context patterns.
These patterns are then recomposed to rapidly iterate agent designs, which are systematically evaluated on controlled robotic planning benchmarks, forming a closed-loop workflow for principled agent improvement.
}
\label{figure:banner}
\end{figure}

\begin{figure}[htb]
\centering
\includegraphics[width=0.7\textwidth]{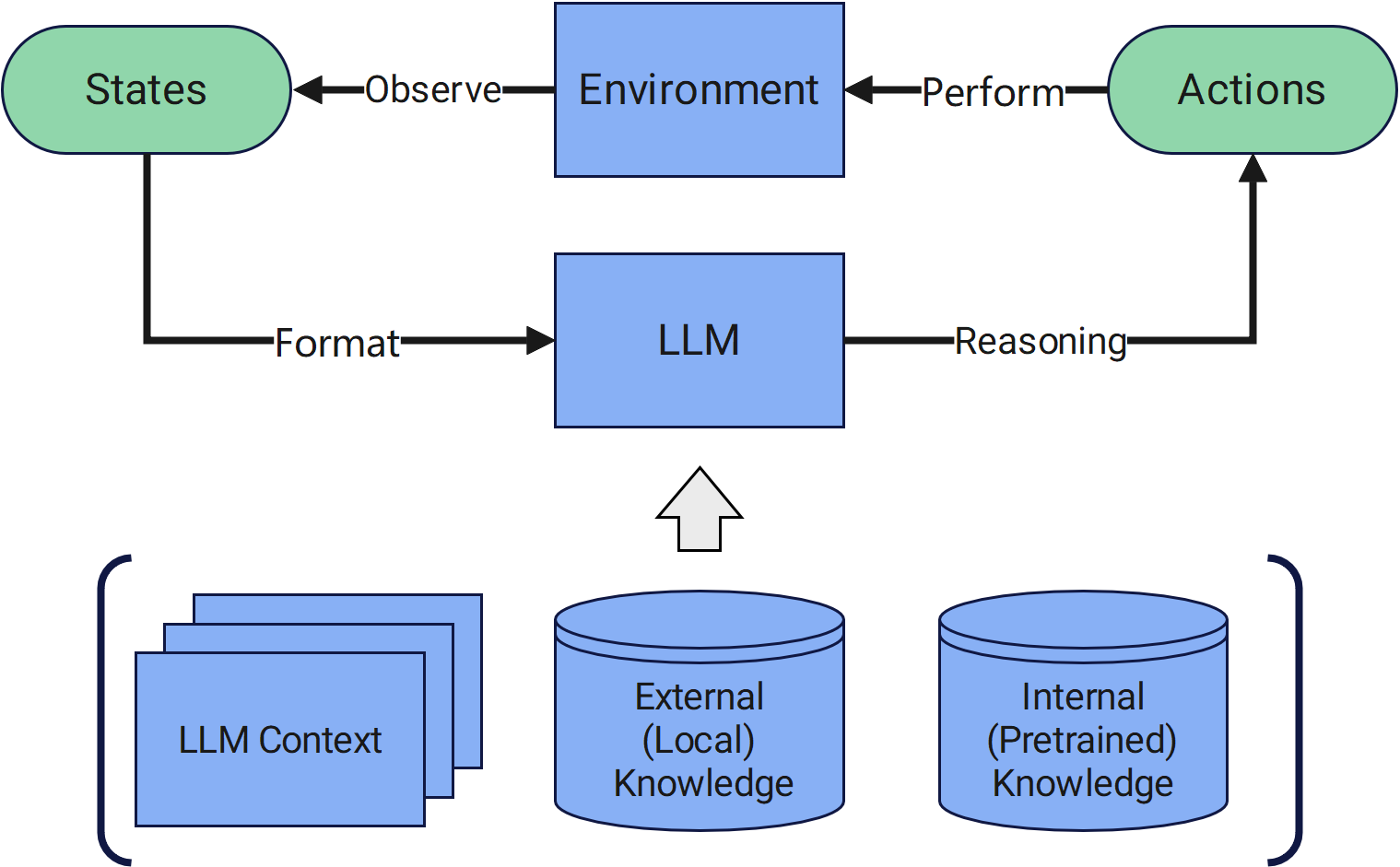}
\caption{The looping architecture of reason–act agents consists of three stages:
(1) the system observes the environment state and formats the observations as requests to the LLM;
(2) the LLM reasons about the next action based on the current environment state, the LLM context, and the relevant knowledge;
(3) the system executes the selected action and waits for the subsequent observations.}
\label{figure:ReActArchitecture}
\end{figure}

In recent years, LLM-based agents have become a prominent research topic in robot task planning, primarily because they are not constrained by the limitations faced by classic symbolic planners such as \texttt{PDDL}~\cite{classic_1998_PDDL} and \texttt{TLPLAN}~\cite{classic_2000_TLPLAN}: namely static environments, deterministic action outcomes, and complete prior knowledge.
Thus, LLM-powered task agents can achieve relatively robust performance under uncertainty.
As a result, a large body of research on LLM-powered agents has emerged. 
These agents are largely built upon similar loop-based real-time architectures, typically following the \texttt{ReAct}~\cite{yao2023react} paradigm, as shown in Figure~\ref{figure:ReActArchitecture}.

However, behind this flourishing diversity of research lies a profound form of confusion. 
When we implemented the well-known \texttt{ReAct} agent and another task-decomposition agent, \texttt{MLDT}~\cite{wu2024MLDT}, within our agent development framework, we were surprised to find that, despite the stark differences in their descriptions at the paper level, their implementation code differed only marginally. 
Moreover, when the implementation-specific discussions in the \texttt{MLDT} work are set aside, its essential distinction from \texttt{ReAct} primarily reduces to an innovation at the prompt level.

We had previously assumed that these two agents exhibited fundamentally significant differences, an assumption that this finding effectively overturned.
This observation immediately prompted a critical question: 
given the large number of agents proposed in the literature, might many of them exhibit no substantial differences once engineering details are abstracted away?

\begin{figure}[htb]
\centering
\includegraphics[width=0.9\textwidth]{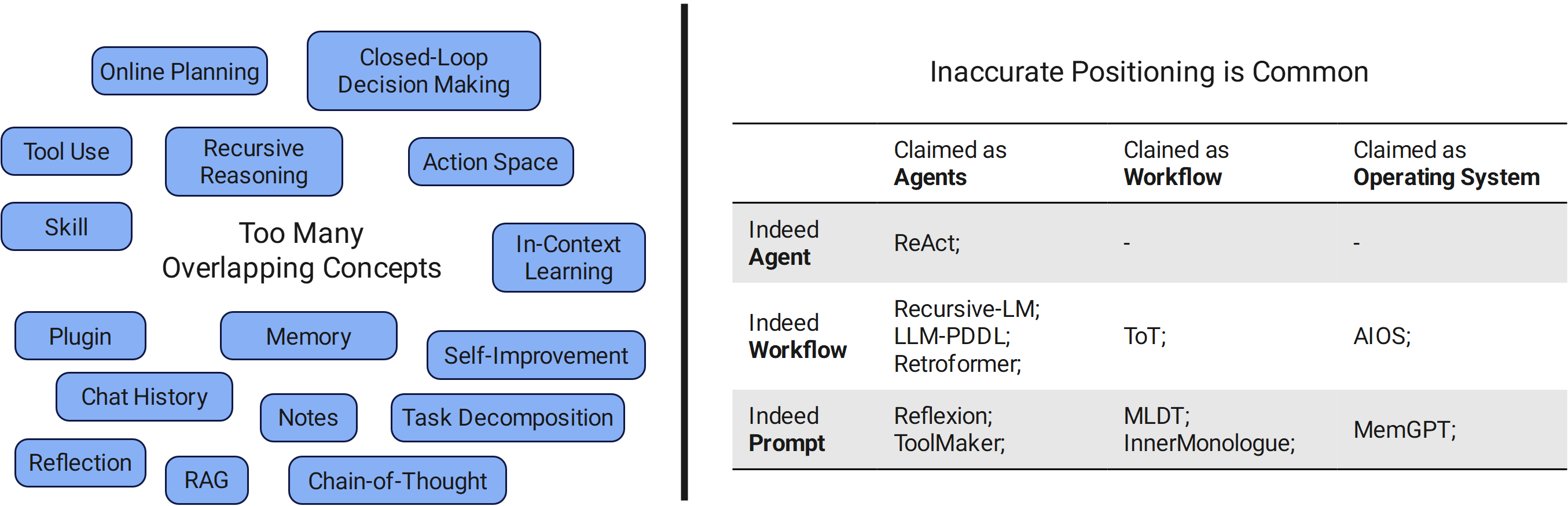}
\caption{
Disorder in current research on LLM-powered agents.
\textbf{Left:} An overabundance of overlapping concepts is currently in use, unnecessarily increasing the cognitive burden on both authors and readers.
\textbf{Right:} The tight coupling between engineering details and methodological descriptions obscures the essence of these innovations, leading to inaccurate positioning of the corresponding research.
Representative works mentioned here include: \texttt{ReAct}~\cite{yao2023react}, \texttt{Recursive-LM}~\cite{zhang2025RecursiveLM}, \texttt{LLM-PDDL}~\cite{silver2022LLM_PDDL}, \texttt{Retroformer}~\cite{yao2023retroformer}, \texttt{Reflexion}~\cite{shinn2023reflexion}, \texttt{ToolMaker}~\cite{cai2023ToolMakers}, \texttt{ToT}~\cite{yao2023TreeOfThoughts}, \texttt{MLDT}~\cite{wu2024MLDT}, \texttt{InnerMonologue}~\cite{huang2022InnerMonologue}, \texttt{AIOS}~\cite{mei2024aios}, \texttt{MemGPT}~\cite{packer2023memgpt}.
}
\label{figure:AcademicDisorder}
\end{figure}

Following this line of inquiry, we conducted a preliminary investigation and found that this phenomenon is not an isolated coincidence but rather a widespread pattern, as intuitively shown in Figure~\ref{figure:AcademicDisorder}.
Current research on LLM-based agents exhibits a considerable degree of disorder.
On the one hand, a large number of overlapping concepts coexist in the literature, placing unnecessary cognitive burdens on both authors and readers.
On the other hand, implementation details are tightly coupled with methodological descriptions, obscuring essential yet incremental innovations behind the appearance of seemingly novel contributions.

We perform this classification based on the modification level of the components introduced relative to \texttt{ReAct}, as well as the impact of these additions on the overall action trajectories of the agent.
For example, if a work can be realized solely by introducing a new prompt on top of \texttt{ReAct}, its contribution is categorized as prompt-level.
This classification is \textbf{NOT} intended to serve as a definitive judgment of the contributions of these works, but rather as an analytical lens to examine whether inaccurate categorization is prevalent in current research on LLM agents.
Our findings indicate that such inaccurate positioning is indeed widespread.

We anticipate that some readers may hold strong objections to this classification. However, such reactions further support our claim that the positioning of innovations in the current literature remains fragmented, underscoring the need for a shared understanding within the research community.

Based on extensive experience in the development of LLM-powered robotic agents, we have gained a direct understanding of the underlying causes of this disorder.
This situation does not stem from subjective factors on the part of researchers, but rather from the exceptionally high engineering complexity of current LLM-based agents.

At present, developers have also not yet reached a consensus on engineering practices in agent development, in contrast to the maturity observed in traditional software engineering. 
As a result, discussions of agent design in the literature inevitably intertwine with implementation-specific considerations. 
Consequently, the engineering complexity of agent systems leaks into and contaminates academic discourse, mystifying the most essential innovations, and increasing the difficulty of meaningful comparison across different agent designs.
This situation has far-reaching consequences for the agent research community. As a feedback effect, the disorder in academic descriptions further causes innovations in agent research to be reused largely at a conceptual or inspirational level, rather than in a concrete and systematic manner. This, in turn, hinders the effective transfer of research outcomes into development practice, forming a negative feedback loop.

In light of this pressing need, this work seeks to draw the community's attention to these issues and to provide an initial attempt to address them.

This paper is a `4-in-1' work, including a model, a methodology, a simulator for benchmarking, and agent designs; each of them can be used independently.
In Section~\ref{section:structural_context_model}, we introduce a formal model for describing, analyzing, and comparing agent designs from the perspective of context structure.
In Section~\ref{section:semantic_dynamics_analysis}, building upon this formal model, we present a methodology for analyzing agent structures and extracting reusable context patterns.
In Section~\ref{section:case_study}, we propose a benchmark for the comprehensive evaluation of task agents, demonstrate how the proposed methodology can be used to study and improve agents, and introduce new agent designs.
While each component is independently applicable, their combination forms an efficient and sustainable closed-loop workflow for agent research and development, in which analysis, pattern discovery, agent design, and benchmarking mutually reinforce each other, as illustrated in Figure~\ref{figure:banner}.

The feasibility of these theoretical methods is enabled by substantial engineering efforts, through which we implemented an automated interoperability layer between LLMs and autonomous systems, as well as a framework built upon the Structural Context Model that significantly improves the reusability of agent components and thereby enhances development efficiency.
This paper does not elaborate on these engineering practices; all source code has been made publicly available.


\section{Structural Context Model for Agent Description}
\label{section:structural_context_model}

In this section, we introduce a theoretical model consisting of a set of symbols and related theories that represent LLM contexts as sequences of pattern functions.
We further elaborate on how to use these symbols to model and explain current engineering practices such as retrieval augmented generation and multi-agent collaboration, to demonstrate its ability to unify these concepts.
Finally, we briefly introduce our programming framework that implements this theoretical model.

\subsection{Relationships with Automata Theory \& Compositional Semantics}

Historically, automata theory and the Turing machine model have served as foundational formalisms for reasoning about software systems.
In a similar vein, it is natural to introduce a formal language, analogous to those used in automata theory, to describe and reason about agent designs in a systematic and implementation-independent manner.

However, LLM-based agents exhibit substantial behavioral differences from the Turing machine model.
First, an LLM does not operate over an explicitly defined and closed set of action symbols.
Second, unlike a Turing machine that reads and executes one symbol at a time, an LLM processes and acts upon the entire context as a whole at each inference step.
In summary, LLM does not treat the context as a deterministic sequence of instructions;
instead, it interprets the semantic meaning of the entire context as a single holistic instruction.

To formally model the `tape of instructions' of an LLM agent, it is necessary to leverage concepts and insights from compositional semantics.
According to compositional semantics, the semantic interpretation of an expression is determined by the semantics of its components together with their mode of composition~\cite{pickel2025compositionality}.

Although compositional semantics has been subject to extensive criticism and debate, these concerns do not preclude its use as a modeling framework for the semantics of context.
This is because:
(1) Agent behaviors are not strictly determined by the exact semantics of the context; rather, a controlled degree of semantic loss is acceptable in practice.
(2) Agent development primarily focuses on discovering prompts that induce specific behaviors, or on identifying the functional roles of particular prompts. Consequently, prompts with higher contextuality can always be replaced by alternative prompts that preserve the relevant semantics while exhibiting lower contextuality.

In summary, by viewing an agent as an automaton that executes the semantics of its context, and by modeling contextual semantics through compositional semantics, we establish the Structural Context Model, which describes and analyzes agents from the perspective of context structure.

\subsection{Proposal: Structural Context Model}

\subsubsection{Symbol Definitions}
\label{section:SCM:symbol_definitions}

In this model, a context is modeled as a sequence of instantiated context patterns; a context pattern is a function that returns context items; and a context item is text or multimodal content with a specific authority level.

\begin{description}
\item[Context Item]
A context item is the minimal constituent unit of a context.
It typically corresponds to a fragment of text, but may also represent multimodal content such as images.
Each context item is associated with a specific role, which indicates the author of this context item; the default role is `user'.
We Denote the space of all context items by $\Omega$; a context item is usually represented by a lowercase Greek letter, for example, $\alpha \in \Omega$.

\begin{itemize}
    \item A text fragment is a context item:
    $\alpha = (\text{`Why didn't this code work?'})$ and $\beta = (\text{`It works on my machine'},\;\text{Agent})$.
    \item The space $\Omega$ is closed under concatenation `$\cdot$', and a context item can be defined with other context items:
    $\gamma = \alpha \cdot \beta$.
    The concatenation operator `$\cdot$' can be omitted, as $\gamma = \alpha \beta$.
    \item The symbol $\boldsymbol{\varepsilon}$ is used to denote an empty context item.
    \item The symbol `$\dots$' represents any context item(s). 
    This symbol is typically used when the expression has no requirements or guaranties for the context items.
    \item The repetition of a context item can be denoted as superscripts. For example, $\alpha^3 = \alpha \cdot \alpha \cdot \alpha$
    \item The production operator is used to represent the concatenation of context items. For example, $\prod_{i=1}^{n} \alpha_i = \alpha_1 \alpha_1 \dots \alpha_n$.
    Unlike in mathematics, the order of expanded elements of the product operator matters because the concatenation of context items is not guaranteed to satisfy the commutative law.
\end{itemize}

\item[Context Pattern] 
A context pattern is a function that returns context items.
In this model, context patterns are the central objects of analysis.
They are denoted in uppercase Latin letters and may have parameters and internal states denoted in lowercase Latin letters.\

\begin{itemize}
    \item For example, a context pattern $A$ with parameters $a,b$ and an internal state $c$ is defined as $A(\,a,b\;|\;c\,) =\alpha \cdot a \cdot c \cdot \beta $.
    \item A context pattern can be defined with another context pattern. 
    For example, $B(\,a,b\;|\;d\,) = \alpha \cdot d \cdot A(\,a,b\,)$
    \item A context pattern instantiated with concrete parameters is a context item. 
    For example, for a pattern $C(\,a, b\,) = a \cdot \alpha \beta \cdot b$, its instantiation $C(\,1, \gamma\,) = \text{`1'} \cdot \alpha \beta \gamma$ is a concatenation of context items.
    \item For parameter-free patterns, the empty item $\boldsymbol{\varepsilon}$ can be used as a parameter to explicitly distinguish a parameter-free pattern itself from its instantiation. For example, this convention distinguishes the instantiation $A(\,\boldsymbol{\varepsilon}\,)$ from the parameter-free pattern function $A(\,)$.
    \item If both the input and the output of a pattern are sequences of context items, the pattern is referred to as a transform pattern, which is a special class of context patterns.
    Transforms can be represented in a simplified notation.
    For example, a transform pattern $D$ that replaces only the leading Context Items of a sequence can be written as $\alpha \dots \xRightarrow{D} \beta \gamma \dots$.
\end{itemize}

\item[Session] 
A session is a tuple of input items sent to the agent and the output items received from the agent, denoted as $\mathcal{S} = (\mathcal{I}, \mathcal{O})$.
The sequence of input items of the $i$-th session $\mathcal{S}_i$ is denoted as $\mathcal{I}_i$; correspondingly, the sequence of output items is denoted as $\mathcal{O}_i$.
According to the concatenation rule of Context Items, both the input and the output of a session are context items.
Therefore, a session can be regarded as an ordered pair of context items.

\item[Activity] 
An activity is an ordered list of sessions that are performed for specific purposes, denoted $\mathcal{A} := [\mathcal{S}_1, \mathcal{S}_2, \dots, \mathcal{S}_n]$.
Equivalently, an activity is an ordered list of pairs of context items.
\end{description}

It is worth noting that our model does not concern itself with the internal processing of patterns as functions, nor with the mapping between states and actions in the program space and those in the LLM space.
Instead, it focuses exclusively on the structural properties of their inputs and outputs.

This design choice is motivated by two considerations.
First, without such an abstraction, it would be impossible to achieve our original goal of providing an implementation-independent model.
Second, this abstraction is practically justified, as we have implemented an interoperability layer that enables automated bidirectional mappings between states and actions in the program space and their counterparts in the LLM space, thereby eliminating the need to explicitly model these details.

\subsubsection{Special Functions}
\label{section:SCM:special_functions}

With the symbols defined in Section~\ref{section:SCM:symbol_definitions}, contexts and patterns can be described in a straightforward manner.
However, additional functions are required to capture the dynamic processes of agent activities.
These functions are denoted by bold uppercase Latin letters.

\begin{description}
    \item[Session Function, $\mathbf{S(\mathcal{I}_i)} \rightarrow \mathcal{O}_i$]
    This function takes a context (a context item, to be more specifically) as the input, passes this context to the agent, and returns the response context from the agent. According to the definition, it is a transform pattern.
    \item[Result Function, $\mathbf{R}_{a,b}(\,\alpha,\beta, \dots \,) \rightarrow a,b$]
    This function parses the response context into program variables.
    In essence, $\mathbf{R}$ acts as the inverse of the context-pattern function: while context patterns map variables (if any) to context items, the result function conversely maps context items back to variables.
    As a usage example, $a = \mathbf{R}(\alpha)$ or $\mathbf{R}_a(\alpha)$ means parse the variable $a$ from the context item $\alpha$; this variable $i$ can be used in patterns after this process.
\end{description}

Many existing studies devote substantial discussion to the interaction between programs and LLMs, as well as to the serialization and deserialization of program states.
In contrast, our model introduces the session function $\mathbf{S}$ and $\mathbf{R}$ to abstract away these engineering details, thereby preventing subsequent analysis from being distracted by such implementation-specific concerns.

As with these two functions, our model delineates a clear boundary between what should be discussed and what should be omitted.
This design represents a direct effort to prevent engineering complexity from further permeating academic research.

\subsection{Unifying Advanced Concepts}

In this subsection, we demonstrate the formal expressions of common components and advanced concepts such as agent memory, retrieval-augmented generation (RAG), and in-context learning within our model.
This serves to substantiate that the proposed model achieves its intended goal, namely, significantly reducing the number of overlapping concepts in existing agent research.

\subsubsection{Agent Memory \& Chat History}
\label{section:SCM:agent_memory}

Agent memories, also known as chat histories, refer to all context items across different sessions in an agent activity.
For an agent activity $\mathcal{A} = [\mathcal{S}_1, \mathcal{S}_2, \dots , \mathcal{S}_n]$,
an agent memory is a pattern $M$ defined in Equation~\ref{equation:agent_memory}.

\begin{equation}
\label{equation:agent_memory}
    M(\;|\;\mathcal{A}) = \mathcal{I}_1 \cdot \mathcal{O}_1 \cdot \mathcal{I}_2 \cdot \mathcal{O}_2 \dots \mathcal{I}_n \cdot \mathcal{O}_n
\end{equation}

The context pattern of an ordinary chatbot with chat history, $F_{\text{chatbot}}(\, \alpha_i \,)$, which has a parameter $\alpha_i$ as text written by the user in the $i$-th turn, can be defined as Equation~\ref{equation:chatbot}

\begin{equation}
\label{equation:chatbot}
\begin{aligned}
    F_{\text{chatbot}}(\, \alpha_i \,) & = M(\boldsymbol{\varepsilon}\;|\;\mathcal{A}) \cdot \alpha_i \\
    & = \mathcal{I}_1 \cdot \mathcal{O}_1 \cdot \mathcal{I}_2 \cdot \mathcal{O}_2 \dots \mathcal{I}_{i-1} \cdot \mathcal{O}_{i-1} \cdot \alpha_i
\end{aligned}
\end{equation}

Moreover, we can present an expression of the response text $\beta_i$ generated by the LLM as Equation~\ref{equation:chatbot_response}.

\begin{equation}
\label{equation:chatbot_response}
\begin{aligned}
    \beta_i & = \mathbf{S} \bigl( F_{\text{chatbot}} \left( \alpha_i \right) \bigr) \\
    & = \mathbf{S}\bigl( M\left(\boldsymbol{\varepsilon}\;|\;\mathcal{A}\right) \cdot \alpha_i\bigr) \\
    & = \mathbf{S}(\mathcal{I}_1 \cdot \mathcal{O}_1 \cdot \mathcal{I}_2 \cdot \mathcal{O}_2 \dots \mathcal{I}_{i-1} \cdot \mathcal{O}_{i-1} \cdot \alpha_i)
\end{aligned}
\end{equation}

\subsubsection{Agent Action \& Chat Tool}
\label{section:SCM:agent_action}

Agent actions are interfaces to program functions for LLMs to use.
Generally, an agent action consists of a program function, textual information about this function and its parameters, and a set of serializers for converting parameters and the return value between textual representations in context space and values in program space.
An agent action, $F_{\text{action}}$, which proxies a program function $A_{\text{action}}(p) \rightarrow r$, can be defined as Equation~\ref{equation:agent_action}.
In this equation, $\mathbf{R}^{-1}$ is the inverse function of the result function $\mathbf{R}$.
$\mathbf{R}$ parses program variables from context items, and $\mathbf{R}^{-1}$ conversely serializes program variables into context items.

\begin{equation}
\label{equation:agent_action}
    F_{\text{action}}(\alpha) = \mathbf{R^{-1}}\Bigl(A_{\text{action}} \bigl(\mathbf{R}(\alpha)\bigr) \Bigr)
\end{equation}

In practice, when an LLM invokes external tools, the results produced by these tools are typically appended to the context.
This side effect of tool invocation under an agent action $F_{\text{action}}$ can be described by Equation~\ref{equation:agent_action_side_effect}.

\begin{equation}
\label{equation:agent_action_side_effect}
     \cdots \xRightarrow{F_{\text{action}}} \cdots F_{\text{action}}(\,\alpha\,) 
\end{equation}

\subsubsection{In-context Learning}
\label{section:SCM:in_context_learning}

The essence of in-context learning (ICL) is injecting and replacing examples within the context.

A piece of an example is denoted as a tuple consisting of a user message in the form of a question $\beta_{\text{Q}}$ and an agent message that provides a verifiable answer $\beta_{\text{A}}$; the $i$-th example in an example buffer is denoted as $(\beta_{(\text{Q}, i)}, \beta_{(\text{A},i)})$.
The example buffer can be denoted as a set of examples: $\mathbb{E} = \left\{(\beta_{(\text{Q}, i)}, \beta_{(\text{A},i)}) | i = 1, \dots, n \right\}$.
Given the question $\alpha$, an in-context learning pattern $F_{\text{ICL}}$ can be defined as Equation~\ref{equation:in_context_learning}

\begin{equation}
\label{equation:in_context_learning}
    F_{\text{ICL}}(\alpha|\mathbb{E}) = \beta_{(\text{Q},1)} \cdot \beta_{(\text{A},1)} \cdot \beta_{(\text{Q},2)} \cdot \alpha_{(\text{A},2)} \dots  \beta_{(\text{Q},n)} \cdot \beta_{(\text{A},n)} \cdot \alpha
\end{equation}

Consider a function $F_{\text{correct}}(\alpha_{\text{Q}})$ that, for any given query parameter $\alpha_{\text{Q}}$, produces the correct answer to the associated question. 
The algorithm for updating the example buffer $\mathbb{E}$ can be defined as Algorithm~\ref{algorithm:in_context_learning}.

\begin{algorithm}
\caption{Updating the Example Buffer}
\label{algorithm:in_context_learning}
\begin{algorithmic}
    \If{$\alpha_{(\text{A}, i)} = F_{\text{correct}}(\alpha_{(\text{Q}, i}))$}
        \State Randomly pick an example $e_j$ from the example buffer $\mathbb{E}$.
        \State $e_j \leftarrow (\alpha_{(\text{Q}, i)}, \alpha_{(\text{A}, i)})$
    \Else
        \State $ \mathbb{E} \leftarrow \mathbb{E} \cup \left\{(\alpha_{(\text{Q}, i)}, F_{\text{correct}}(\alpha_{(\text{Q}, i})))\right\} $
    \EndIf
\end{algorithmic}
\end{algorithm}

Alternatively to the updating algorithm Algorithm~\ref{algorithm:in_context_learning}, the example buffer can also be considered a set containing all historical tuples of questions and answers, and a custom importance function is deployed to rank these examples by their ``educational value''; for example, examples for which the LLM has previously output the incorrect answer have higher chances of being selected.

From the Equation~\ref{equation:in_context_learning}, we can see that it has the same structure as the Equation~\ref{equation:agent_memory} of the agent memory, except that the collections have different semantic meanings and updating algorithms.
This may encourage the belief that agent memories and ICL probably rely on the same fundamental mechanism, and that ICL works because LLMs possess an internal drive to reproduce their own prior behaviors.

This example also demonstrates the power of the Structural Context Model; it provides a rigorous method to identify the similarities and differences in current engineering practices from a structural perspective.

\subsubsection{Retrieval Augmented Generation}
\label{section:SCM:retrieval_augemented_generation}

A retrieval augmented generation (RAG) pattern takes text as input and returns related supplementary materials as output.
Typically, it uses cosine similarity of embedding vectors to search for the most related supplementary materials.

A RAG pattern can be provided in two different manners:
1) provided as an agent tool (as explained in Section~\ref{section:SCM:agent_action});
2) provided as a transform pattern that receives query text as the parameter and returns the retrieved supplementary materials.
Here, we present the definition for the latter one as $F_{\text{RAG}}$ in Equation~\ref{equation:retrieval_augmented_generation}.
In this equation, 
the query text is denoted as $\alpha$;
the $i$-th supplementary material is defined as $\beta_i$;
the knowledge database is denoted as $\mathbb{B} = \{ \beta_1, \dots, \beta_n \}$, which is a set of supplementary materials.

\begin{equation}
\label{equation:retrieval_augmented_generation}
    F_{\text{RAG}}(\alpha) = \alpha \cdot (\text{`Supplementary material:'}) \cdot \mathop{\arg\max}\limits_{\beta_i} \{ \mathop{cossim}(\alpha, \beta_i) | \beta_i \in \mathbb{B} \}
\end{equation}

Comparing Equation~\ref{equation:in_context_learning} of ICL and Equation~\ref{equation:retrieval_augmented_generation} of RAG, it is obvious that both ICL and RAG are injecting data into the context; the fundamental difference lies in the data source and the way this data is selected: ICL selects examples from the example buffer based on their importance, while RAG selects supplementary materials from the knowledge database based on their cosine similarities with the query text.

\subsubsection{Multi-agent Collaboration}
\label{section:SCM:multi_agents}

In current engineering practice, the main purpose of multi-agent collaboration is to decompose a complex task into several relatively easier ones and delegate them to sub-agents specifically customized for heterogeneous tasks.

Since the meaning of sub-agents is to generate results of decomposed sub-tasks for the main agent to use, they can be modeled as context patterns.

Equation~\ref{equation:multi_agent_collaboration} is an example of a compositional agent $F_{\text{main}}$ that answers questions about physics and chemistry.
It has four sub-agents, $F_{\text{physics}}$, $F_{\text{chemistry}}$, $F_{\text{RAG}}$, and $F_{\text{router}}$. 
Sub-agents $F_{\text{physics}}$ and $F_{\text{chemistry}}$ are agents customized for their corresponding domains, with specially designed prompts;
sub-agent $F_{\text{RAG}}$ is the RAG pattern defined in Section~\ref{section:SCM:retrieval_augemented_generation};
sub-agent $F_{\text{router}}(\alpha) \rightarrow \{0, 1\}$ is the pattern that determines whether the query text $\alpha$ is more related to physics (when it returns $0$) or chemistry (when it returns $1$).

\begin{equation}
\label{equation:multi_agent_collaboration}
\begin{aligned}
F_{\text{router}}(\alpha) = {}& \mathbf{R}_i\Bigl(\mathbf{S}\bigl( (\text{`Output 0 if this question is primarily about physics, '}) \cdot \\ 
&{} (\text{`and 1 if it is primarily about chemistry.'}) \cdot \alpha \bigr)\Bigr), \quad i \in \{0, 1\} \\
F_{\text{main}}(\alpha) = \;& 
\begin{cases}
    F_{physics}\bigl( F_{RAG}(\alpha) \bigr), & i = 0 \\
    F_{chemistry}\bigl( F_{RAG}(\alpha) \bigr), & i = 1 \\
\end{cases}
\end{aligned}
\end{equation}

\subsection{Comparing Agents at a Fundamental Level}
\label{section:SCM:comparing_agents}
By abstracting away the execution and implementation-level engineering details of agents, the \texttt{Structural Context Model} enables analysis and comparison of agent designs and behaviors from a more fundamental perspective.

In this subsection, we demonstrate this key capability by comparing \texttt{ReAct}-based and \texttt{MLDT}-based agents through their formal expressions under the \texttt{Structural Context Model}.

Denote the problem description and rules as $\alpha$, action lists as $\beta$.
For the $i$-th step, denote the current status as $\gamma_i$.
Denote the agent memory pattern as $M(\;|\;\mathcal{A})$ from Section~\ref{section:SCM:agent_memory}.

The context patterns corresponding to reasoning and to the \texttt{ReAct}-based agent can be formally defined as shown in Equation~\ref{equation:SCM:ReAct}.
Similarly, the context patterns of the task-decomposed MLDT agent can be given as Equation~\ref{equation:SCM:MLDT}.
The primary differences between the two agents are highlighted in \textcolor{red}{red}.

\begin{equation}
\label{equation:SCM:ReAct}
\begin{gathered}
    D_{\text{reasoning}}(\gamma_i) = \mathbf{S}\bigl(\alpha\beta \cdot \gamma_i \cdot (\textcolor{red}{\text{``Analyze the current situation...''}}\bigr)
    \rightarrow \textcolor{red}{r_i} \\
    \begin{aligned}
        F_{\text{ReAct}}(\gamma_i) & = \alpha \beta \cdot M(\varepsilon \;|\;\mathcal{A}) \cdot \prod_{i} \gamma_i \cdot (\text{``Perform an action.''}) \cdot \textcolor{red}{D_{\text{reasoning}}(\gamma_i)} \\
        & = \alpha \beta \cdot M(\varepsilon \;|\;\mathcal{A}) \cdot \prod_{i} \gamma_i \cdot (\text{``Perform an action.''}) \cdot \textcolor{red}{r_i}
    \end{aligned}
\end{gathered}
\end{equation}

\begin{equation}
\label{equation:SCM:MLDT}
\begin{gathered}
    \begin{aligned}
        D_{\text{MLDT}}(\gamma_i) & = \mathbf{S}\bigl(\alpha\beta \cdot \gamma_i \cdot (\textcolor{red}{\text{``Break down the task into three hierarchies...''}}\bigr) \\
        & \rightarrow \textcolor{red}{\mathcal{P}_i = [\, p_1, p_2, ..., p_n \,]}
    \end{aligned} \\
    \begin{aligned}
        F_{\text{MLDT}}(\gamma_i) & = \alpha \beta \cdot M(\varepsilon \;|\;\mathcal{A}) \cdot \textcolor{red}{D_{\text{MLDT}}(\gamma_0)} \cdot \prod_{i} \gamma_i \cdot (\text{``Perform an action.''}) \\
        & = \alpha \beta \cdot M(\varepsilon \;|\;\mathcal{A}) \cdot \textcolor{red}{\mathcal{P}_0} \cdot \prod_{i} \gamma_i \cdot (\text{``Perform an action.''})
    \end{aligned}
\end{gathered}
\end{equation}

From the formal expressions of the two agents, their primary differences can be identified both intuitively and straightforwardly.
Specifically, the key distinction lies in the prompt-level roles of the information injected into the agent's context prior to action execution, namely, \textit{reasoning} versus \textit{planning}.

In addition, the \texttt{ReAct} agent performs a reasoning step before each action is taken.
In contrast, the original formulation of \texttt{MLDT} does not specify conditions or mechanisms for plan adjustment or updating; as a result, it can be interpreted as generating a static plan based on the initial state $\gamma_0$.

Given that \texttt{ReAct} was proposed in 2022 and \texttt{MLDT} in 2024, and considering both the substantial temporal gap and their influence, we along with other researchers, initially expected the two approaches to exhibit significant structural differences.
However, the analysis and comparison conducted under the \texttt{Structural Context Model} yield a markedly different conclusion.
This further demonstrates the strong capability of the model to reveal the essential nature of agent design.

\subsection{Engineering Efforts}

Building upon the proposed model, we have implemented an open-source framework, \texttt{ContextCompose}, which enables rapid construction and iterative refinement of agent designs through the composition of reusable context patterns.
Inspired by declarative UI frameworks, \texttt{ContextCompose} introduces the notion of composing into the domain of agent design, whereby new context patterns are designed and described through the composition of existing ones.

This paradigm substantially increases the modularity and reusability of agent software artifacts, thereby significantly accelerating both development and research iteration cycles.

This paper focuses primarily on the theoretical contributions of the proposed model; the innovations of the framework from the perspective of agent software engineering will be elaborated in separate publications.


\section{Semantic Dynamics Analysis for Pattern Discovery}
\label{section:semantic_dynamics_analysis}

In this section, we present \texttt{Semantic Dynamics Analysis}, a method for analyzing the functional roles and relative contributions of different context parts.
By characterizing how context parts influence agent behavior, this method facilitates the identification and abstraction of reusable context patterns from existing agent designs.

Beyond its analytical utility, \texttt{Semantic Dynamics Analysis} also highlights a key distinction between our \texttt{Structural Context Model} and existing approaches.
Rather than focusing solely on whether specific improvements derived from a method are effective, we additionally emphasize the sustainability of the method itself, that is, whether it enables readers to continuously and independently discover new improvement opportunities.

\subsection{Theoretical Roots of Dynamic Semantics}
\label{section:SDA:dynamic_semantics}

The core idea of Dynamic Semantics is that it considers the meaning of a sentence as its `potential' to update the context~\cite{nouwen2022dynamic_semantics}.
Under classic static semantics, such as compositional semantics, knowing the meaning of a sentence amounts to knowing the conditions under which it is true.
In contrast, Dynamic Semantics defines meaning in terms of how a sentence transforms the current information state.

Dynamic Semantics is particularly well suited to our setting, where context is analyzed through its incremental construction and the resulting changes in context-level metrics.
Concretely, the analysis begins by resetting the context to an empty state, and then progressively reconstructing it by reintroducing context components in their original order, while observing the induced semantic dynamics.
This controlled reconstruction process enables us to identify the functional roles of individual components as well as the magnitude of their influence.

\subsection{Definition and Indicators}
\label{section:SDA:definition_and_indicators}

\begin{figure}[htb]
\centering
\includegraphics[width=0.9\textwidth]{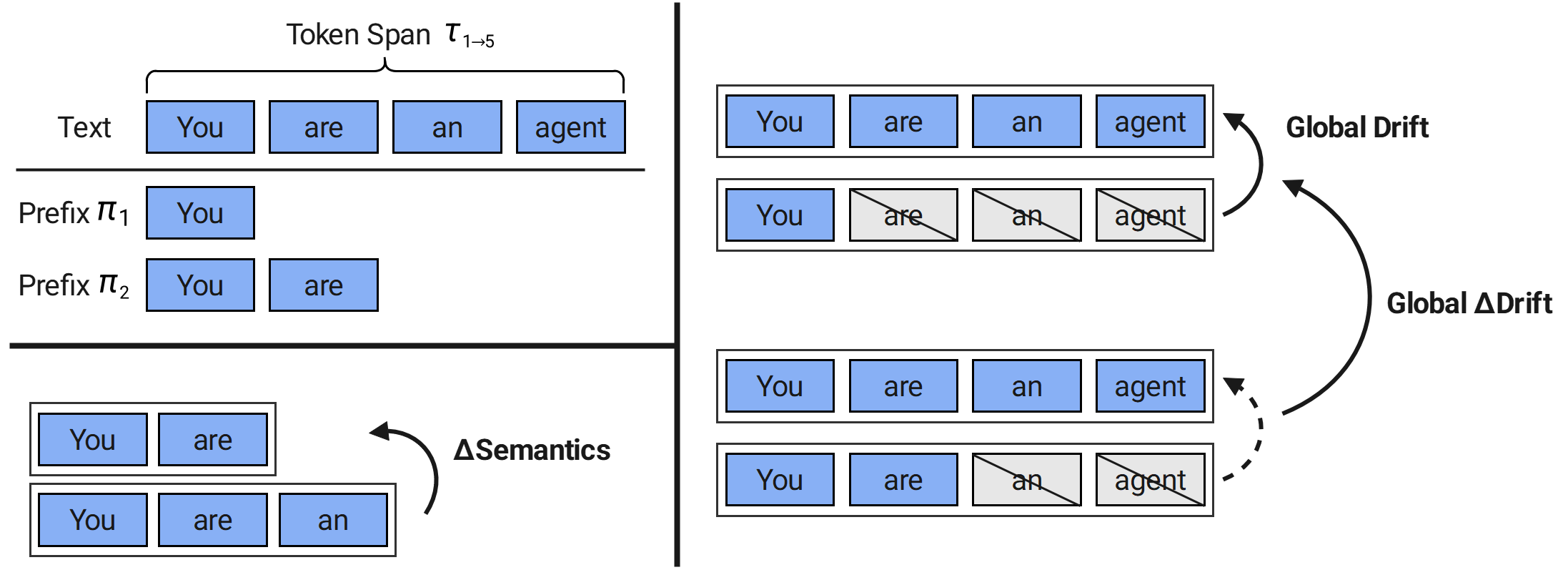}
\caption{Intuitive description of key terms in \texttt{Semantic Dynamics Analysis}.
$\Delta$Semantics measures the semantic distance between successive prefixes, capturing the semantic update potential introduced by the appended token.
Global Drift measures the semantic distance between a prefix and the full text, indicating the extent to which the overall meaning would differ if the text were to end at the current prefix.
Global $\Delta$Drift measures the rate at which the current prefix approaches the final meaning of the full text, reflecting the velocity of semantic convergence.
}
\label{figure:SDA:Terms}
\end{figure}

In this section, we explain several technical terms used in this analytical method and the proposed indicators.
Figure~\ref{figure:SDA:Terms} provides an intuitive illustration of the main terms and indicators used in \texttt{Semantic Dynamics Analysis}.

\begin{description}
    \item[Token] 
    A token is the minimal unit for LLMs (the Transformer architecture~\cite{vaswani2017attention}, to be more specific) to process.
    Tokens are acquired from text by a tokenizer with a static dictionary that maps text parts to a specific integer identifier (token).
    The $i$-th token is denoted as $\tau_i$;
    naturally, for any text, the $0$-th token is empty: 
    $\tau_0 = \boldsymbol{\varepsilon}$.
    
    \item[Prefix Text]
    Prefix texts are cumulative partial texts built from the token sequence, e.g., the first prefix is token 1, the second is tokens 1 to 2, and so on.
    For a text 
    $\gamma = \tau_1 \cdot \tau_2 \dots \tau_n$, the $i$-th prefix is denoted as 
    $\pi_i := \tau_1 \cdot \tau_2 \dots \tau_i$;
    naturally, the $0$-th prefix is empty: 
    $\pi_0 = \boldsymbol{\varepsilon}$,
    and the $n$-th prefix is the text itself: 
    $\pi_n = \gamma$.
    Meanwhile, according to the definition that texts are also context items in Section~\ref{section:SCM:symbol_definitions}, tokens, token spans, and prefix texts are all elements of the context item space $\Omega$.
    
    \item[Semantics]
    The semantics of a text $\alpha$ is defined as a set $\mathbb{S}_\alpha$ of context-updating potentials.
    The notion of ``context'' here does not refer to a LLM context but follows Dynamic Semantics that is introduced in Section~\ref{section:SDA:dynamic_semantics}.
    For two texts $\alpha$ and $\beta$, we write $\mathbb{S}_\alpha \supseteq \mathbb{S}_\beta$ to indicate that $\alpha$ semantically subsumes $\beta$.

    \item[Embedding Vector \& Semantic Centroid]
    An embedding vector is a high-dimensional numerical representation generated by an embedding model from an input text.
    Embedding vectors are typically normalized to lie on a unit hyper-sphere in the embedding space, which allows semantic relatedness between texts to be measured via geometric distances.
    For a given text (or token span), its embedding vector is constructed such that the distance to embedding vectors of semantically more related texts is smaller than the distance to those of less related texts, under the same representation model.
    \begin{itemize}
        \item In this work, an embedding vector is treated as a representation-valued functional of the semantic set $\mathbb{S}$ associated with a token span.
        The embedding vector reflects the \textbf{semantic centroid} of the token span in the embedding space,  providing a measurable geometric proxy for its underlying semantic structure.
        \item Moreover, for a token span with complex semantic content, its semantic centroid can be viewed as a composition of multiple relatively independent \textbf{semantic sub-centroids}, each corresponding to a coherent semantic component whose mutual relatedness is comparatively low.
    \end{itemize}
    
    \item[Semantic Distance]
    Semantic distance quantifies the degree of semantic divergence between two texts in an embedding space, reflecting how dissimilar their meanings are under a given representation model.
    It is defined as a geometric measure between the embedding vectors associated with the texts.
    Denoting the embedding vector of a text $\alpha$ by $\bm{v}_{\alpha}$, the semantic distance between two texts $\alpha$ and $\beta$ is measured by the cosine similarity of their embedding vectors, as shown in Equation~\ref{equation:CA:semantic_distance}:
    \begin{equation}
    \label{equation:CA:semantic_distance}
    \begin{aligned}
    \operatorname{semdist}(\alpha, \beta) := & 1 - \operatorname{cossim} (\bm{v}_\alpha, \bm{v}_\beta) \\
    = & 1 - \frac{\bm{v}_\alpha \cdot \bm{v}_\beta}{\lVert \bm{v}_\alpha \rVert \lVert \bm{v}_\beta \rVert}
    \end{aligned}
    \end{equation}
\end{description}

Here, we adapt embedding vectors as the measurable estimation of semantics, rather than the internal states of LLM encoders.
Due to the attention mechanism~\cite{vaswani2017attention} of LLMs, it is internal states generated by the LLM encoder that directly determine the outputs of the whole LLM.
However, due to multiple concerns, directly observing internal states is not the most suitable approach.
The first concern is that internal states can only be acquired from self-hosted open-source models.
Adapting this indicator will constrain this analytical method to a limited range of scenarios.
The second concern is that the values of internal states vary according to different model architectures and even different versions of the same model.
It is almost impossible to define indicators and theorems that are stable across models based on these unstable representations.
The third concern is that these internal states are sensitive to fine-tunings, especially fine-tuning over specific instruction keywords.
This property makes it unsuitable for revealing generally applicable properties and rules of context patterns.
Therefore, we adapt embedding vectors as indirect estimation of LLM internal states.

\begin{description}
    \item[$\Delta$Semantics]
    The $\Delta$Semantics of a token is defined as its semantic contribution to a prefix, measured by the semantic distance between the embeddings of relevant prefixes. \\
    \begin{itemize}
    \item For a contiguous token span 
    $\tau_{i \rightarrow j} := \tau_i \cdot \tau_{i+1} \dots \tau_j$, 
    its $\Delta$Semantics with respect to the original prefix context is defined as 
    $\Delta S(\tau_{i \rightarrow j}) := \operatorname{semdist}(\pi_j, \pi_{i-1})$. \\
    The token-level $\Delta$Semantics is a special case when $i = j$, i.e., 
    $\Delta S(\tau_i) = \Delta S(\tau_{i \rightarrow i}) = \operatorname{semdist}(\pi_i, \pi_{i-1})$.

    \item More generally, when the $\Delta$Semantics of a token span $\tau_{i \rightarrow j}$ is measured relative to an external base text $\gamma$, the base text must be explicitly specified:
    $\Delta S_{\gamma}(\tau_{i \rightarrow j}) = \operatorname{semdist}\bigl(\gamma \cdot (\tau_i \cdot \tau_{i+1} \dots \tau_j), \gamma\bigr)$

    \item By convention, the first token has $\Delta$Semantics equal to $0$, since no prior prefix exists for comparison. \\
    This measure reflects the relative importance of a token within the prefixes that contain it.
    \end{itemize}
    
    \item[Global Drift]
    The semantic distance between each prefix and the full-text (i.e., the final prefix), reflecting how far each prefix text is from the complete meaning: 
    ${D(\pi_i) = \operatorname{semdist}(\pi_i, \pi_n)}$.
    Intuitively, the Global Drift provides a geometric approximation of the distance from the prefix to the semantic centroid of the text in the embedding space.

    \item[Global $\Delta$Drift]
    The Global $\Delta$Drift is defined as the change in Global Drift between successive prefix texts:
    $\Delta D(\pi_i) = D(\pi_{i-1}) - D(\pi_{i})$
    \begin{itemize}
        \item This definition is chosen such that a positive value of $\Delta D(\pi_i)$ indicates that the prefix moves closer to the final semantic state, while a negative value indicates divergence.
        \item  By convention, the first token has a value of $0$ since there is no earlier prefix.
        \item From an interpretative perspective, the Global $\Delta$Drift quantifies the contribution of the current token to the semantic progression of its associated prefix.
        A larger Global $\Delta$Drift suggests that the token induces a substantial shift in the prefix semantics.
        In particular, pronounced peaks in Global $\Delta$Drift often signal the onset of a transition from one semantic sub-centroid to another, thereby marking a boundary between the respective influence ranges of different semantic sub-centroids.
    \end{itemize} 
\end{description}

\begin{figure}[htb]
    \centering
    \includegraphics[width=0.9\textwidth]{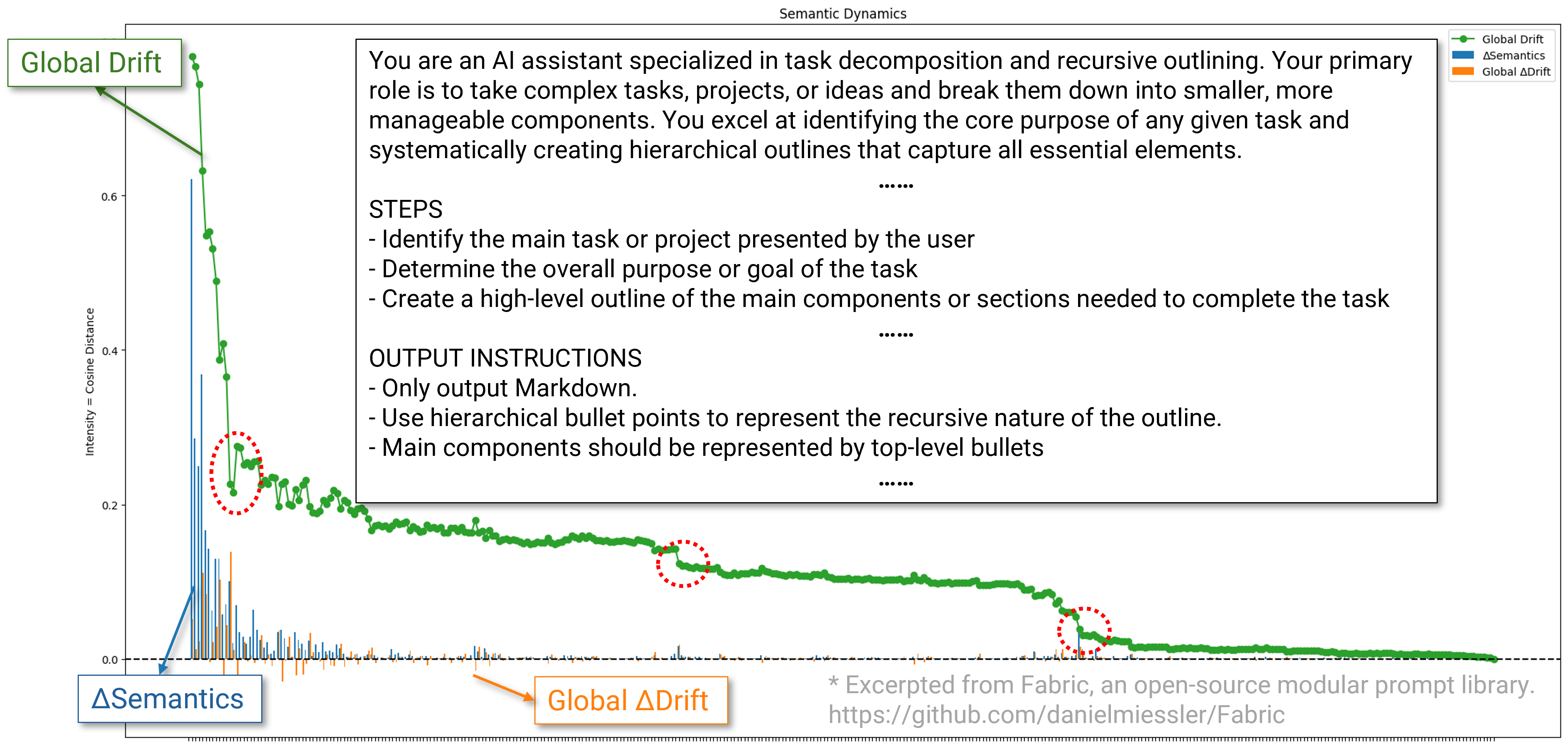}
    \caption{
    Results of \texttt{Semantic Dynamics Analysis} for a task-decomposition prompt excerpted from an open-source prompt library.
    The blue line represents Global Drift, the blue bars indicate $\Delta$Semantics, and the orange bars denote Global $\Delta$Drift.
    Red circles mark potential semantic separation points, where Global $\Delta$Drift exhibits relatively large values.
    }
    \label{figure:SDA:task_decomposition_example}
\end{figure}

To provide an intuitive understanding of the components of \texttt{Semantic Dynamics Analysis}, Figure~\ref{figure:SDA:task_decomposition_example} illustrates the analysis results for a task-decomposition prompt excerpted from an open-source prompt library.

\begin{figure}[htb]
    \centering
    \includegraphics[width=1.0\textwidth]{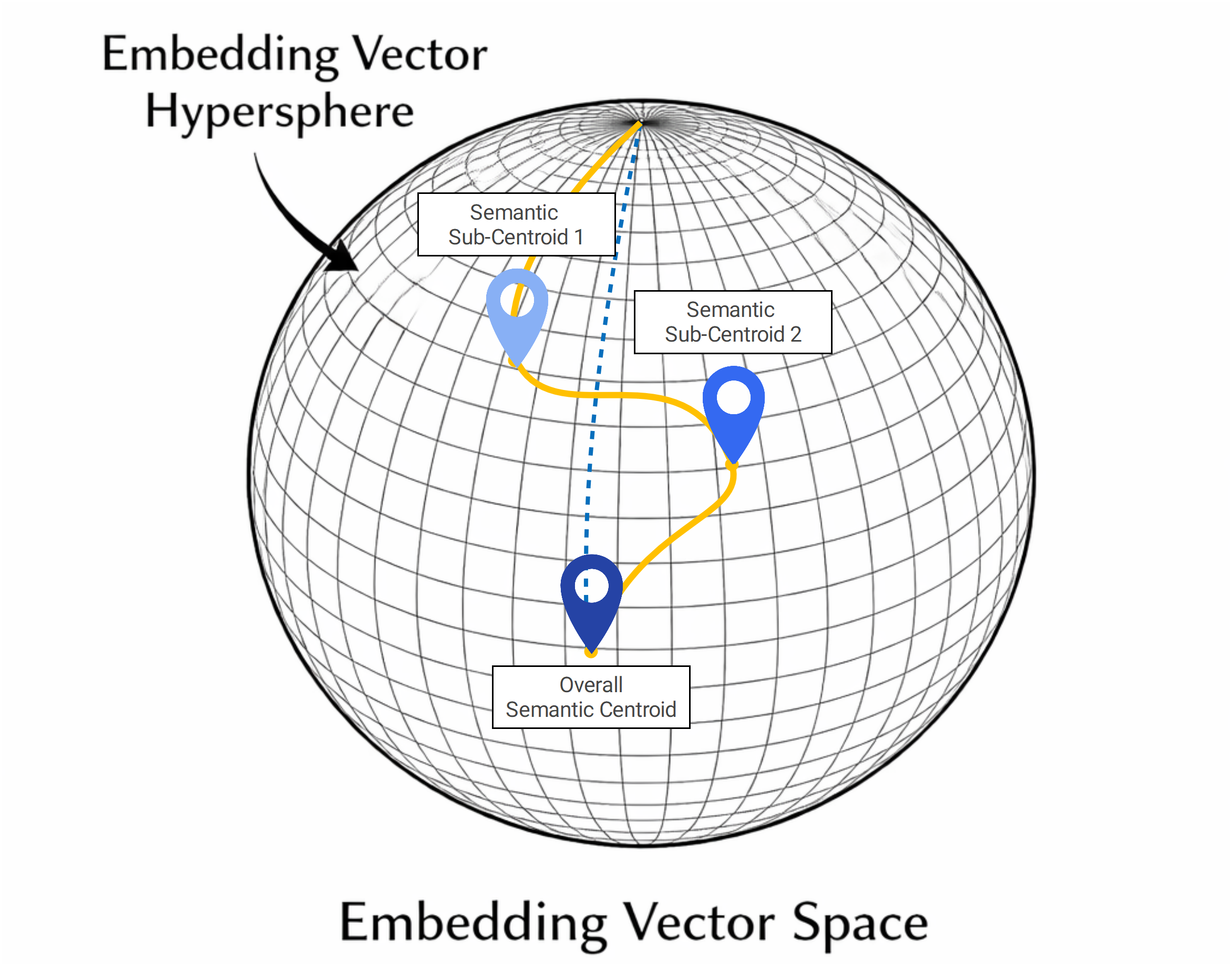}
    \caption{A schematic illustration showing how semantic sub-centroids influence the Global $\Delta$Drift.
    The first semantic sub-centroid is marked by a light-blue waypoint, the second semantic sub-centroid by a blue waypoint, and the semantic centroid of the full text by a dark-blue waypoint.
    As each token is appended, the current prefix moves along the actual transition trace, shown as a solid orange curve.
    Due to the presence of semantic sub-centroids, the rate at which the prefix approaches the overall semantic centroid (which is the Global $\Delta$Drift) varies over time, rather than following the optimal transition trace indicated by the blue dashed curve, which always possess the maximum Global $\Delta$Drift.
    The sudden significant change of Global $\Delta$Drift indicate the direction of current prefix approaching the overall semantic centroid has dramatically changed, and thus suggests that the current prefix is semantically close to a semantic sub-centroid.
    }
    \label{figure:SDA:semantic_transition}
\end{figure}

When semantic analysis is applied to identify natural separation points in an existing context structure, Global $\Delta$Drift captures the rate at which each prefix approaches the overall semantic centroid of the full text.
Variations in this rate indirectly indicate the presence of semantic sub-centroids along the context trajectory.
Figure~\ref{figure:SDA:semantic_transition} provides an intuitive illustration of how semantic sub-centroids influence Global $\Delta$Drift.
The first and second semantic sub-centroids are indicated by light-blue and blue waypoints, respectively, while the semantic centroid of the full text is marked by a dark-blue waypoint.
As tokens are appended, the current prefix evolves along the actual semantic transition trace, shown as a solid orange curve.
Owing to the presence of semantic sub-centroids, the rate at which the prefix approaches the overall semantic centroid, that is, Global $\Delta$Drift, varies over time, rather than following the optimal transition trace depicted by the blue dashed curve, which consistently attains the maximum Global $\Delta$Drift.
A sudden and significant change in Global $\Delta$Drift therefore indicates a sharp shift in the direction of semantic convergence, suggesting that the current prefix lies at a potential semantic sub-centroid.

In summary, Global $\Delta$Drift is used to identify semantic segmentation points within the context structure.
$\Delta$Semantics characterizes the relative importance of individual tokens within a segment.
The overall trend of Global Drift over a segment reflects the contribution of that segment to the entire context structure.

\subsection{Geometric Structure of Context Pattern Semantics}

With the indicators introduced in Section~\ref{section:SDA:definition_and_indicators}, we are able to analyze the semantic properties and functional roles of context patterns.

\subsubsection{Semantic Properties and Relationships}
\label{section:SDA:geometric_properties}

When studying the properties of (instantiated) context patterns in a context, we mainly focus on the following special properties and relationships:

\begin{description}
    \item[Semantic Inclusion]
    For two token spans $\alpha$ and $\beta$, we say $\alpha$ semantically includes $\beta$ if: $\mathbb{S}_{\alpha \cdot \beta} = \mathbb{S}_{\beta \cdot \alpha} = \mathbb{S}_{\alpha}$.
    From an engineering perspective, if a pattern $\alpha$ semantically includes $\beta$, then appending $\beta$ to a context that already contains $\alpha$ is expected to have no additional effect on the resulting context semantics.    
    
    \item[Semantic Parallelism]
    For two token spans $\alpha$ and $\beta$, we say $\alpha$ and $\beta$ are semantically parallel if: $\mathbb{S}_\alpha = \mathbb{S}_\beta$, i.e., both $\mathbb{S}_\alpha \subset \mathbb{S}_\beta$ and $\mathbb{S}_\alpha \supset \mathbb{S}_\beta$ hold.
    Semantic parallelism is transitive: if $\alpha$ is semantically parallel to $\beta$ and $\beta$ is semantically parallel to $\gamma$, then $\alpha$ is semantically parallel to $\gamma$.
    From an engineering perspective, semantically parallel patterns are interchangeable: using any one or any combination of them is expected to produce the same effect on the context.
    
    \item[Semantic Orthogonality]
    For two texts $\alpha$ and $\beta$, we say $\alpha$ and $\beta$ are semantically orthogonal if: $\mathbb{S}_\alpha \cap \mathbb{S}_\beta = \varnothing$.
    In this case, neither $\alpha$ nor $\beta$ semantically includes the other.
    Moreover, the concatenation of semantically orthogonal texts is order-invariant.
    From an engineering perspective, removing any pattern from a set of orthogonal patterns in a context is expected to cause a substantial change in the resulting semantics.

    \item[Semantic Order-Invariance]
    For two token spans $\alpha$ and $\beta$, we say the combination set $\{\alpha, \beta\}$ is semantically order-invariant if: $\mathbb{S}_{\alpha \cdot \beta} = \mathbb{S}_{\beta \cdot \alpha}$.
    From an engineering perspective, the relative order of order-invariant combinations does not affect the context semantics.
    
    \item[Semantic Idempotence]
    A token span $\alpha$ is semantically idempotent if for any positive integer $k > 1$, it can be repeated for $k$ times in the context and have the same effect on the context semantics:
    $\mathbb{S}_\alpha = \mathbb{S}_{\alpha^k}$
    where 
    $\alpha^k := \underbrace{\alpha \cdot \alpha \cdots \alpha}_{k\ \text{times}}$.
    \begin{itemize}
        \item It can be proven that a token span is semantically idempotent if and only if it is semantically absorbable by itself.
        \item From an engineering perspective, semantically idempotent patterns instantiated with identical parameters need not be included multiple times, as repeated occurrences can be safely reduced to a single instance to save tokens.
        Conversely, when different patterns inevitably contain the same semantically idempotent sub-patterns, no special handling is required for such repetitions, since they do not introduce additional semantic effects.
    \end{itemize}
    
\end{description}

\subsubsection{Bridging Semantic Properties and Measurable Indicators}
\label{section:SDA:geometric_propostions}

Based on the indicators defined in Section~\ref{section:SDA:definition_and_indicators} and the properties established above, we state the following propositions: 
Proposition~\ref{proposition:CA:inclusion_sub_additivity}, 
Proposition~\ref{proposition:CA:orthogonality_super_additivity},
Proposition~\ref{proposition:CA:delta_semantics_idempotence}
, and Proposition~\ref{proposition:CA:delta_semantics_order_invariance}.

\begin{proposition}[Sub-additivity of $\Delta$S under Semantic Inclusion]
\label{proposition:CA:inclusion_sub_additivity}
Let $\alpha$ and $\beta$ be two token spans.
If $\alpha$ semantically includes $\beta$, then for any base text $\gamma \in \Omega$,
\[
    \Delta S_\gamma(\alpha \cdot \beta)
    \;\le\;
    \Delta S_\gamma(\alpha) \,+\, \Delta S_\gamma(\beta)
\]
\end{proposition}


\begin{proposition}[Super-additivity of $\Delta$S under Semantic Orthogonality]
\label{proposition:CA:orthogonality_super_additivity}
Let $\alpha$ and $\beta$ be two token spans.
If $\alpha$ and $\beta$ are semantically orthogonal, then for any base text $\gamma \in \Omega$
\[
    \Delta S_\gamma(\alpha \cdot \beta)
    \;\ge\;
    \Delta S_\gamma(\alpha) \,+\, \Delta S_\gamma(\beta)
\]
\end{proposition}

\begin{proposition}[Approximate $\Delta$S Idempotence under Semantic Idempotence]
\label{proposition:CA:delta_semantics_idempotence}
Let $\alpha$ be a token span.
If $\alpha$ is semantically idempotent, then for a user-defined tolerance threshold $\eta > 0$ and any base text $\gamma \in \Omega$, 
\[
    \lvert \, \Delta S_\gamma(\alpha^k)
    \,-\,
    \Delta S_\gamma(\alpha) \, \rvert 
    \;<\; \eta
\]
\end{proposition}

\begin{proposition}[Approximate $\Delta$S Invariance under Semantic Order-Invariance]
\label{proposition:CA:delta_semantics_order_invariance}
Let $\alpha$ and $\beta$ be two token spans.
If the combination set of $\{\alpha, \beta\}$ is semantically order-invariant, then for a user-defined tolerance threshold $\eta > 0$ and any base text $\gamma \in \Omega$, 
\[
    \lvert \, \Delta S_\gamma(\alpha \cdot \beta)
    \,-\,
    \Delta S_\gamma(\beta \cdot \alpha) \, \rvert 
    \;<\; \eta
\]
\end{proposition}

These propositions bridge the special semantic properties and relationships with measurable indicators derived from embedding vectors, thereby enabling their discovery through mathematical and algorithmic methods.

Moreover, these propositions also contributes to a shift in context engineering from predominantly experience-driven, heuristic empirical studies toward computable and analyzable data-driven research.

\subsection{Methodology}

In this section, we describe a workflow for discovering reusable patterns from existing prompts, which consists of two main steps: segmentation and parameterization.

We have developed a software tool to analyze a prompt and generate the corresponding analysis charts; however, the level of automation remains limited.
In particular, the parameterization step still requires manual intervention.

The workflow consists of the following steps:

\begin{enumerate}
    \item \textbf{Tokenization.}
    A tokenizer is applied to the prompt to obtain a token sequence, which serves as the input for subsequent analysis.
    
    \item \textbf{Semantic dynamics analysis.}  
    Based on the definitions in Section~\ref{section:SDA:definition_and_indicators}, we compute the $\Delta$Semantics for each token, as well as the Global Drift and Global $\Delta$Drift for each prefix.
    
    \item \textbf{Segmentation via Global $\Delta$Drift.}  
    Prefixes associated with abnormally high Global $\Delta$Drift values are identified.  
    As discussed in Section~\ref{section:SDA:definition_and_indicators}, such anomalies suggest the presence of nearby semantic sub-centroids. 
    The ending tokens of these prefixes are therefore used as semantic separation points, partitioning the prompt into multiple segments.
    
    \item \textbf{Property verification of segments.}  
    The propositions introduced in Section~\ref{section:SDA:geometric_propostions} are applied to examine whether the extracted segments exhibit the semantic properties and relationships described in Section~\ref{section:SDA:geometric_properties}.

    \item \textbf{Segment refinement.}  
    Within each segment, tokens with high $\Delta$Semantics are treated as locally important, whereas tokens with relatively low $\Delta$Semantics are considered less influential and may be removed with limited impact on the local semantics.

    \item \textbf{Parameterization of influential content.}  
    Token spans exhibiting both high $\Delta$Semantics and high Global $\Delta$Drift are regarded as semantically influential at both the segment and prompt levels.  
    Such spans are natural candidates for parameterization, as modifying or replacing them typically induces substantial semantic changes.  
    The final decision of parameterization remains flexible and can be guided by practical engineering requirements.
\end{enumerate}

\subsubsection{Demonstration on a Task Decomposition Agent}
\label{section:SDA:demonstration_task_decomposition}
Text~\ref{text:SDA:task_decomposition_example} shows an example prompt for task decomposition excerpted from an open-source prompt library \textit{Fabric}\footnote{The URL to the repository: \url{https://raw.githubusercontent.com/danielmiessler/Fabric}.}.

In this section, we demonstrate how \texttt{Semantic Dynamics Analysis} can be applied to existing context structures to extract reusable patterns, which are later employed in the our agent designs presented in Section~\ref{section:case_study}.

\begin{tcbtextbox}{Prompt for Task Decomposition}
{halign lower=right, label=text:SDA:task_decomposition_example}
You are an AI assistant specialized in task decomposition and recursive outlining. Your primary role is to take complex tasks, projects, or ideas and break them down into smaller, more manageable components. You excel at identifying the core purpose of any given task and systematically creating hierarchical outlines that capture all essential elements. 
......
STEPS
- Identify the main task or project presented by the user
- Determine the overall purpose or goal of the task
- Create a high-level outline of the main components or sections needed to complete the task
......
OUTPUT INSTRUCTIONS
- Only output Markdown.
- Use hierarchical bullet points to represent the recursive nature of the outline.
- Main components should be represented by top-level bullets
......
\end{tcbtextbox}

\begin{figure}[htb]
\centering
\includegraphics[width=0.9\textwidth]{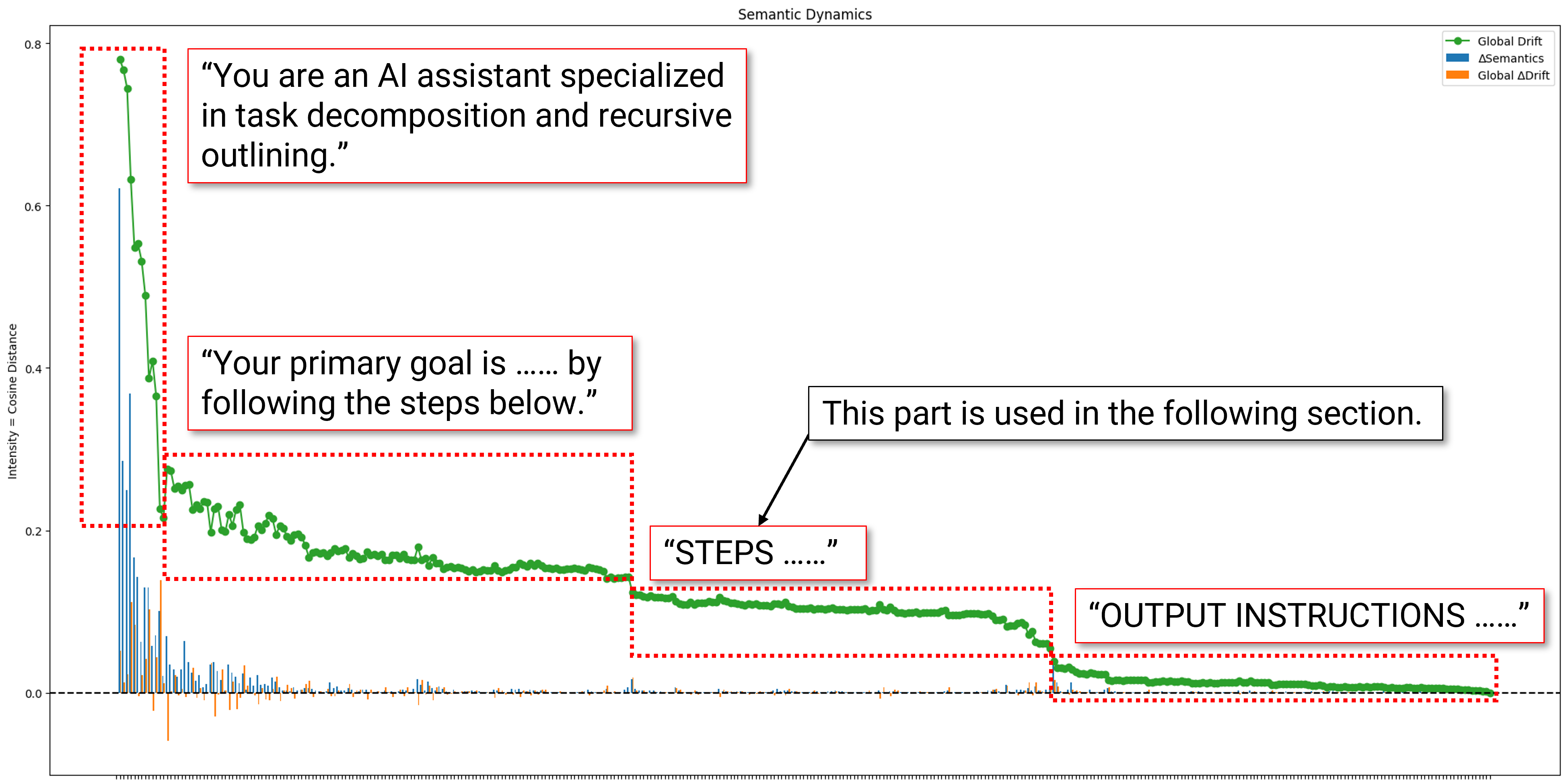}
\caption{
Semantic segments identified from the results of \texttt{Semantic Dynamics Analysis}.
The prompt is partitioned into four segments: role settings, goal settings, detailed steps, and output format settings.
The segment corresponding to detailed steps is reused in the agent designs presented in Section~\ref{section:case_study}.}
\label{figure:SDA:task_decomposition_segments}
\end{figure}

The detailed results of \texttt{Semantic Dynamics Analysis} are provided in Appendix~\ref{appendix:task_decomposition_result}.  
Figure~\ref{figure:SDA:task_decomposition_segments} illustrates the semantic segments identified based on these results.  
According to the analysis, the prompt can be partitioned into four segments: (1) role settings, (2) goal settings, (3) detailed steps, and (4) output format settings.
Moreover, the $\Delta$Semantics values reported in Table~\ref{table:appendix:pairwise_segments_delta_semantics} indicate that segment pairs (1, 2) and (3, 4) exhibit relatively strong semantic order invariance.

Among them, the segment corresponding to detailed steps is reused in the agent designs presented in Section~\ref{section:case_study}.

\section{Case Study}
\label{section:case_study}

In this chapter, we conduct a comprehensive case study to demonstrate the applicability of the methods introduced in Section~\ref{section:structural_context_model} and Section~\ref{section:semantic_dynamics_analysis}. 
Based on these methods, we design three novel agents with different context structures.
In addition, we reproduce several representative baselines for comparison, including the widely adopted \texttt{ReAct} agent, the \texttt{MLDT} agent, and a minimal baseline agent that follows a simple observe–act loop without explicit planning or memory mechanisms.

All 6 agents are evaluated in a suite of 15 task scenarios with varying dynamic factors and difficulty levels. 
Each agent is tested with 100 independent repeated trials per scenario to ensure statistical robustness.

\subsection{Problem Settings}

\begin{figure}[htb]
\centering
\includegraphics[width=0.9\textwidth]{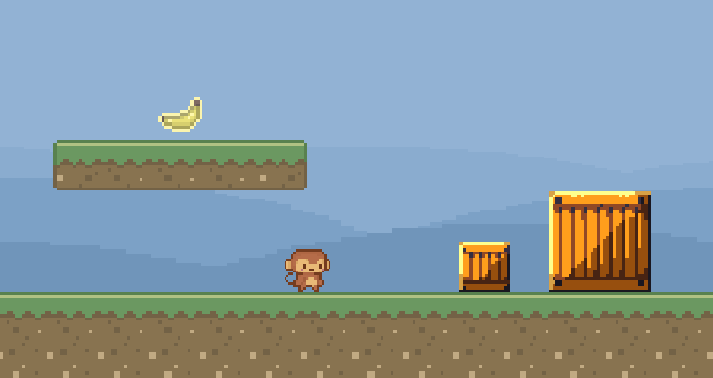}
\caption{
The classic monkey-banana problem. The goal of this problem is to control the monkey, manipulate boxes, and ultimately obtain the banana.
}
\label{figure:CS:classic_monkey_banana_problem}
\end{figure}

The monkey–banana problem~\cite{richmond1966monkey_banana} is a classic benchmark for evaluating robotic task planning.
As illustrated in Figure~\ref{figure:CS:classic_monkey_banana_problem}, the objective of the task is to control the monkey, manipulate boxes, and ultimately obtain the banana.

Building upon the classic monkey–banana problem, we introduce a set of dynamic factors to construct a series of variant monkey–banana problems, making the task setting closer to the real-world challenges faced by modern robotic intelligent systems.
These dynamic factors violate the core assumptions relied upon by classical planners, including a static environment, deterministic action outcomes, and complete prior knowledge.
As a result, such variants cannot be solved within the classical planning paradigm.
This further underscores the importance of LLM-based task agents.

The problem scenarios in our experiments are organized into five major categories: \textit{Classic}, \textit{Dual Bananas}, \textit{Shortsighted Monkey}, \textit{Over-weight Monkey}, and \textit{Comprehensive}.
Each category is further instantiated at three difficulty levels: \textit{easy}, \textit{medium}, and \textit{hard}.
In the remainder of this paper, we refer to these problem instances uniformly as \textit{scenes}, following the terminology used in our simulator.
The elaboration of different difficulty levels are detailed in Section~\ref{section:CS:difficulty_factors}, while the elaboration of different categories are provided in Section~\ref{section:CS:scene_categories}.

\subsubsection{Difficulty Factors}
\label{section:CS:difficulty_factors}

\begin{figure}[htb]
\centering
    \begin{subfigure}[b]{0.3\textwidth}
        \centering
        \includegraphics[width=\textwidth, height=2cm]{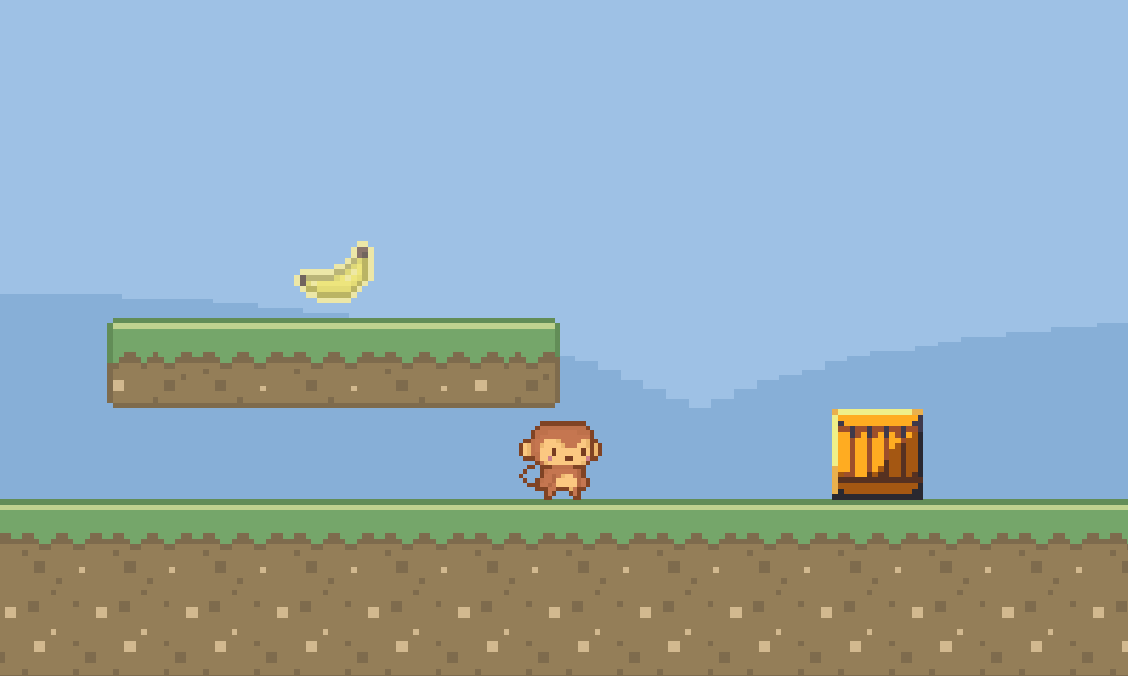}
        \caption{Single Banana: Easy }
    \end{subfigure}
    \quad
    \begin{subfigure}[b]{0.3\textwidth}
        \centering
        \includegraphics[width=\textwidth, height=2cm]{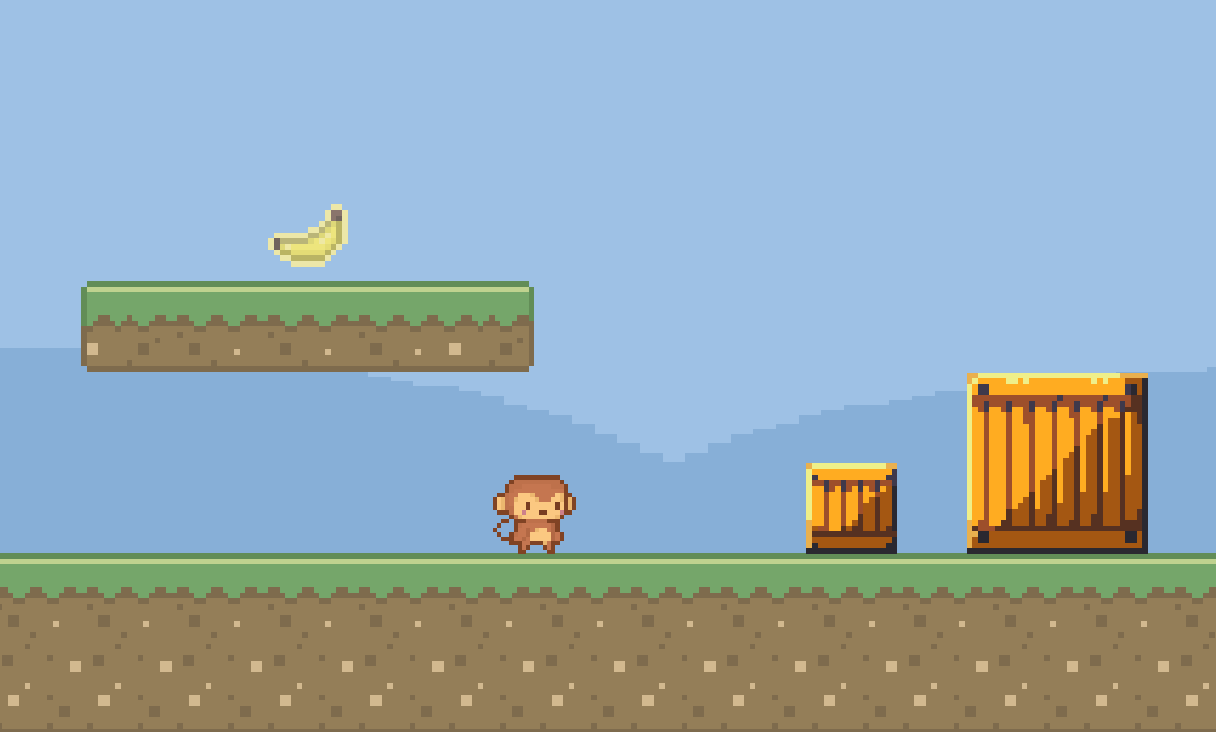}
        \caption{Single Banana: Medium}
    \end{subfigure}
    \quad
    \begin{subfigure}[b]{0.3\textwidth}
        \centering
        \includegraphics[width=\textwidth, height=2cm]{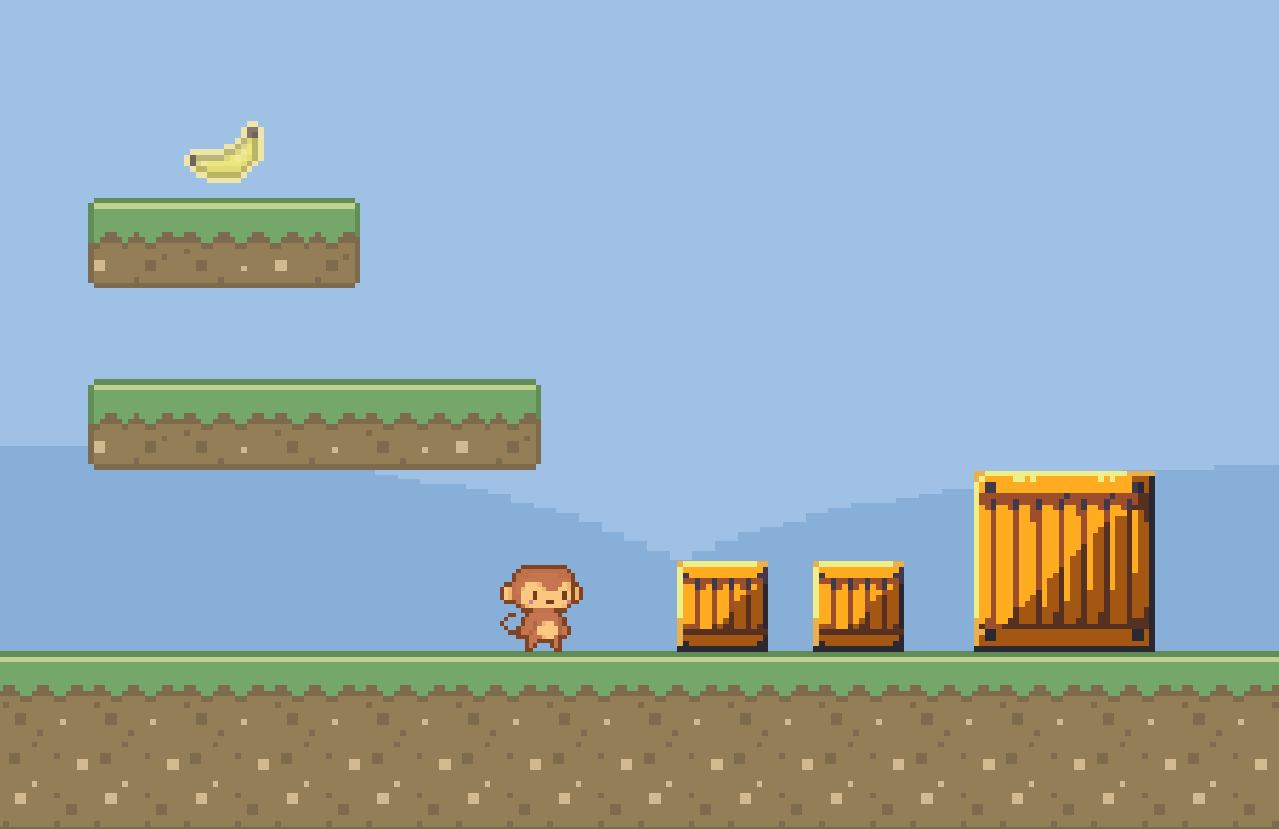}
        \caption{Single Banana: Hard}
    \end{subfigure}
    \quad
    \begin{subfigure}[b]{0.3\textwidth}
        \centering
        \includegraphics[width=\textwidth, height=2cm]{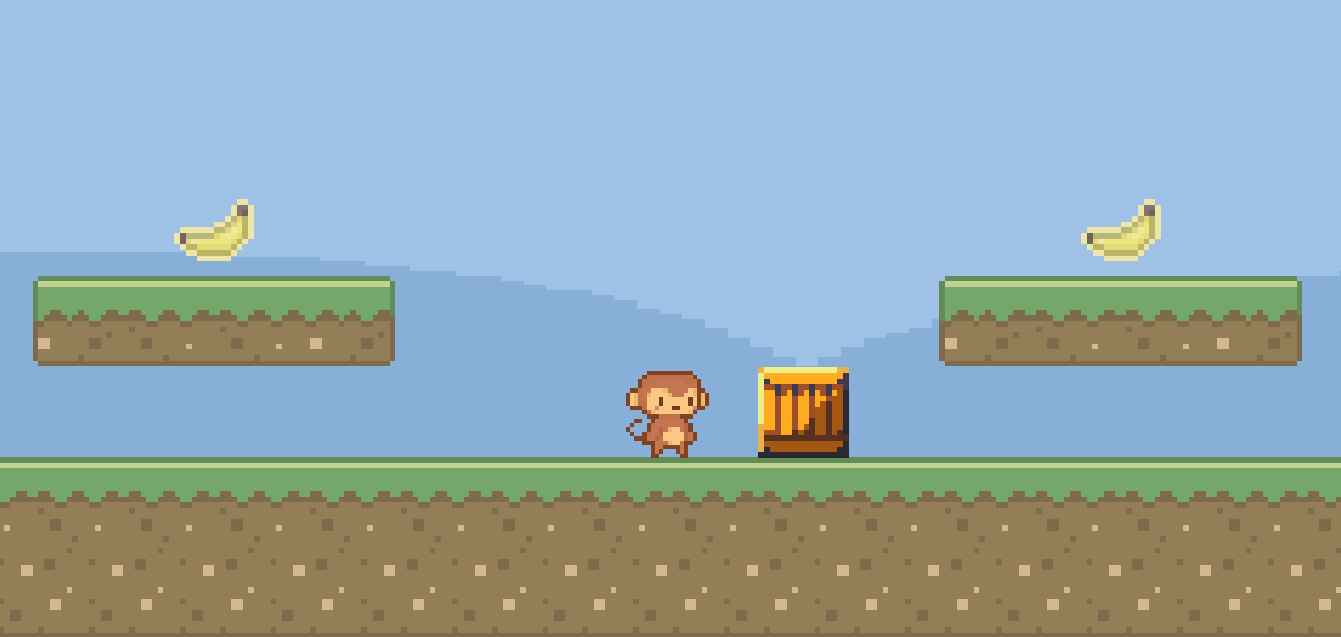}
        \caption{Dual Bananas: Easy}
    \end{subfigure}
    \quad
    \begin{subfigure}[b]{0.3\textwidth}
        \centering
        \includegraphics[width=\textwidth, height=2cm]{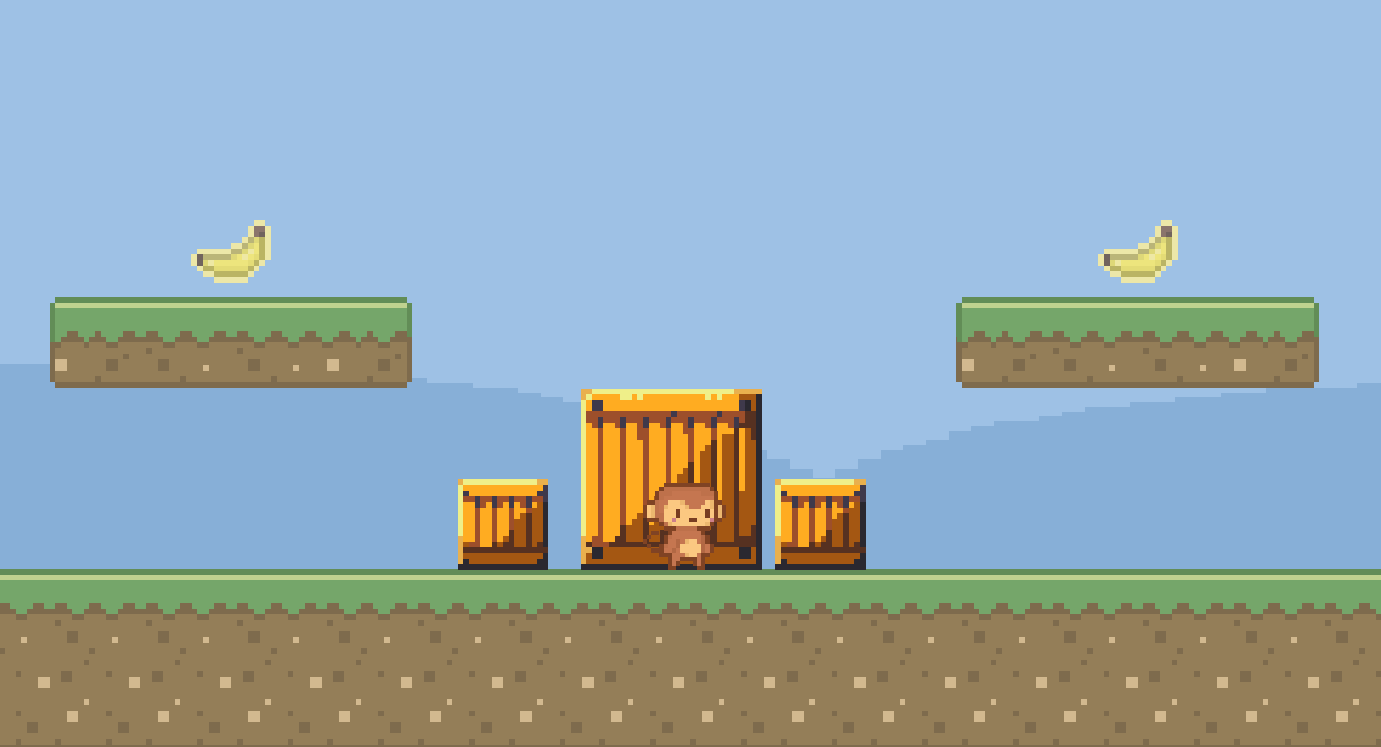}
        \caption{Dual Bananas: Medium}
    \end{subfigure}
    \quad
    \begin{subfigure}[b]{0.3\textwidth}
        \centering
        \includegraphics[width=\textwidth, height=2cm]{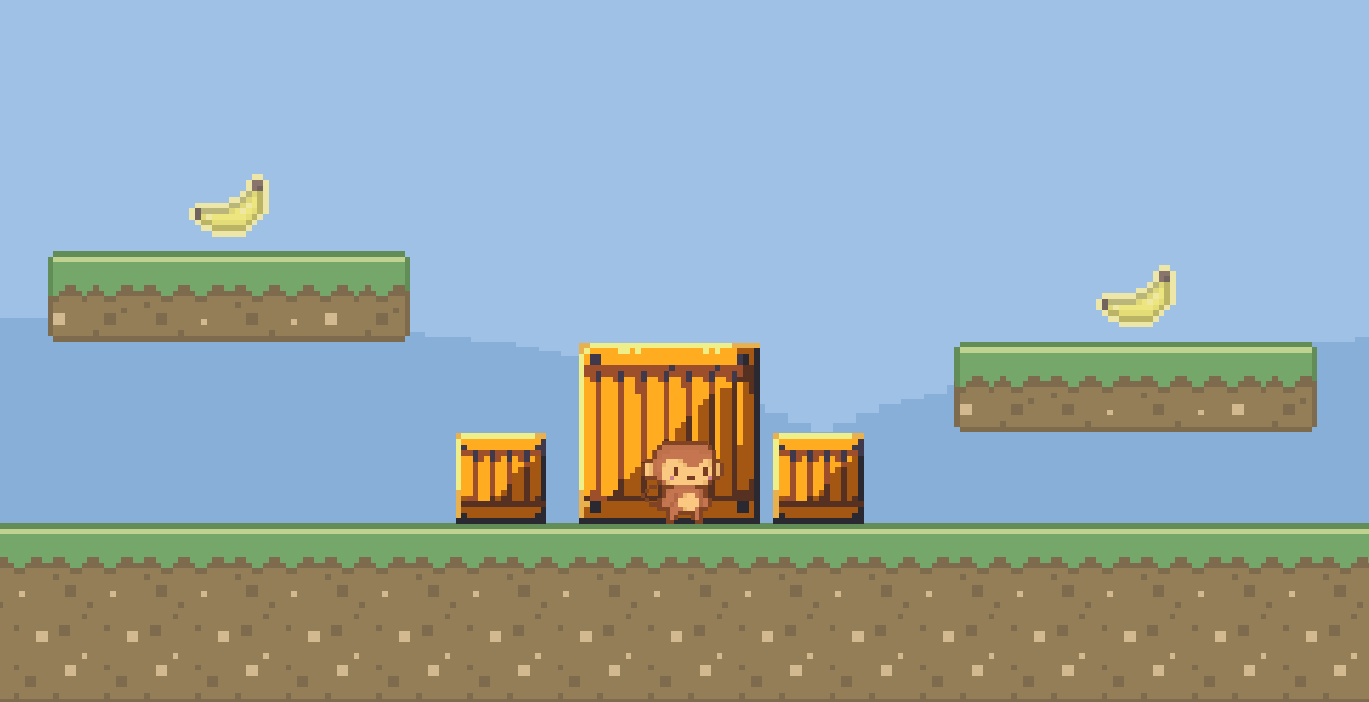}
        \caption{Dual Bananas: Hard}
    \end{subfigure}
    \quad
\caption{Overview of monkey–banana problem scenes with increasing difficulty.
The difficulty levels are determined by both the number of target bananas (single or dual) and the complexity of the environment configuration, resulting in progressively more demanding task planning scenarios.}
\label{figure:CS:problems_overview}
\end{figure}

Scenes with different difficulty levels are illustrated in Figure~\ref{figure:CS:problems_overview}.

In the easy scenes, the environment contains a single small box and a single elevated platform.
This setting is designed to verify whether an agent possesses the basic capability to solve the monkey–banana problem.

In the medium scenes, the environment includes one small box and one large box.
The monkey must sequentially climb the small box and then the large box in order to reach the platform holding the banana.
This setting primarily evaluates the agent's reasoning ability under multi-step action dependencies.

In the hard scenes, the environment contains two platforms arranged at different heights.
To reach the second platform where the banana is placed, the monkey must carry a small box, sequentially climb the small box and the large box, and place the small box onto the first platform.
Beyond further increasing reasoning complexity, this setting also enables evaluation of the agent's foresight, as reflected by the number of steps required to anticipate the need to transport an additional box in advance.

\subsubsection{Scene Categories}
\label{section:CS:scene_categories}

In total, the experimental setup consists of 15 scenes, indexed by Scene IDs from 1 to 15.
These scenes are organized into five distinct categories, with each category comprising three instances corresponding to the different difficulty levels defined in Section~\ref{section:CS:difficulty_factors}.
Within each category, the difficulty increases with the scene index.
For example, Scenes~1$\sim$3 correspond to the classic monkey–banana problem, where Scene~1 represents the easy setting, Scene~2 the medium setting, and Scene~3 the hard setting.

\begin{figure}[htbp]
\centering
\includegraphics[width=0.9\textwidth]{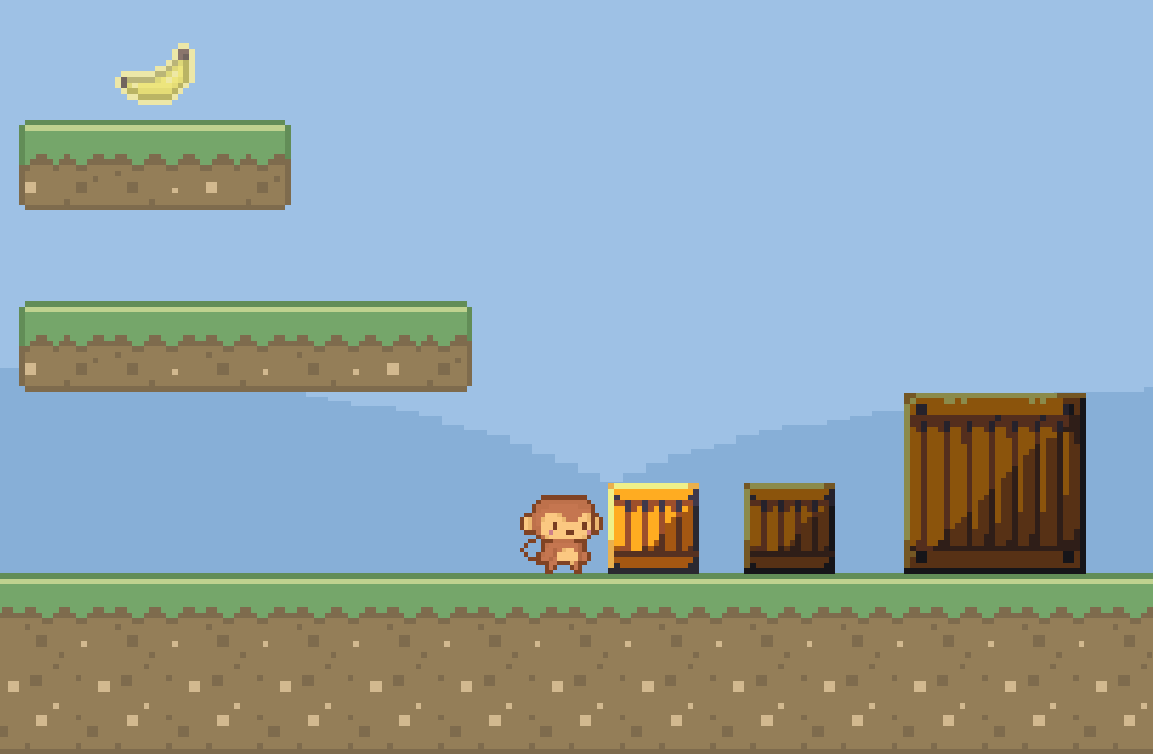}
\caption{Illustration of the `Shortsighted Monkey' mechanism.
When a box moves beyond the monkey's interaction range, its detailed information becomes unavailable.
The dark-colored boxes in the figure indicate objects whose attributes are marked as \texttt{unknown} due to being outside the interaction range.}
\label{figure:CS:variant_problems_shortsighted}
\end{figure}

\begin{figure}[htbp]
\centering
\includegraphics[width=0.7\textwidth]{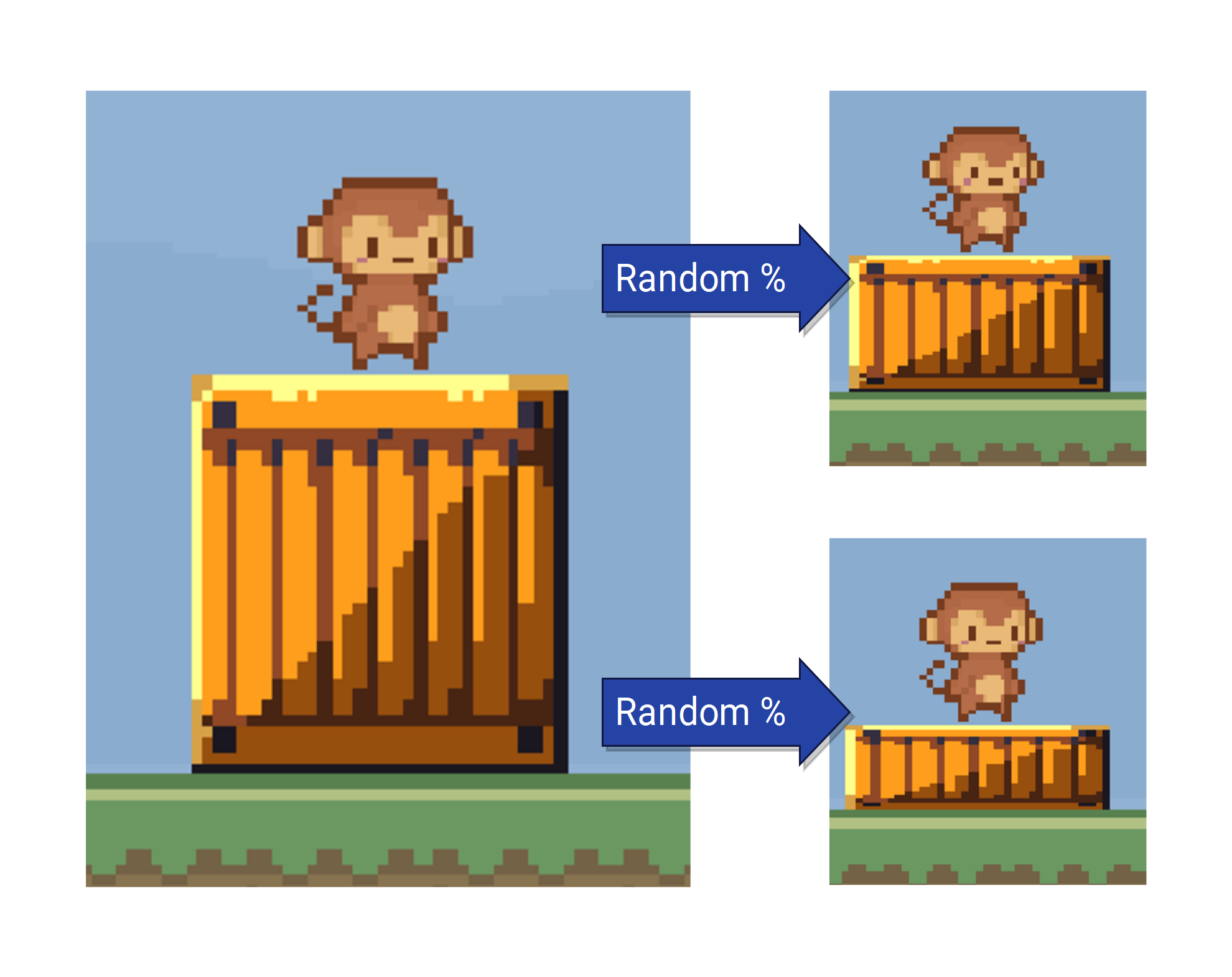}
\caption{Illustration of the `Over-weight Monkey' mechanism: each time the monkey climbs onto a box, there is a probability that the box will be crushed, reducing its height to a random proportion.}
\label{figure:CS:variant_problems_overweighted}
\end{figure}

\begin{description}
    \item[(ID 1$\sim$3) `Classic'] 
    These scenes correspond to the canonical monkey-banana problem instantiated at different difficulty levels.

    \item[(ID 4$\sim6$) `Dual Bananas'] 
    In this category, two bananas are provided within the scene.
    The objective is to evaluate the agent's ability to consistently commit to a single target when multiple equivalent goals are available, rather than exhibiting indecisive or oscillatory behavior.
    Notably, in the hard setting, the increased difficulty is not introduced by adding an additional platform.
    Instead, the two targets incur different achievement costs, with one banana being easier to obtain than the other, thereby testing the agent's ability to flexibly assess and select among competing goals.

    \item[(ID 7$\sim$9) `Shortsighted Monkey']
    This category violates the classical planner assumption of complete predefined knowledge.
    In these scenes, when the distance between the monkey and a box exceeds the interaction range, the simulator no longer provides detailed information about that box, such as its dimensions, and instead marks it as \texttt{unknown}.
    This category is designed to evaluate the agent's memory capability. \\
    Figure~\ref{figure:CS:variant_problems_shortsighted} provides an intuitive illustration of this characteristic.
    In the figure, the gray boxes indicate objects whose detailed information is rendered as \texttt{unknown} because they are beyond the monkey's interaction range.

    \item[(ID 10$\sim$12) `Over-weight Monkey']
    This category violates the classical planner assumption of deterministic action outcomes.
    In these scenes, whenever the monkey climbs onto a box, there is a 50\% probability that the box will be crushed, reducing its height to a random proportion of its original height (ranging from 50\% to 70\%).
    The probability of being crushed differs across boxes.
    As a result, a platform that would normally be reachable by stacking one large box and one small box may instead require an additional small box placed on top of the crushed large box if the latter collapses. \\
    Many agents perform multiple attempts within a scene.
    Such irreversible side effects caused by crushing boxes can invalidate the scene knowledge acquired from previous attempts, thereby further evaluating the agent's ability in integrated analysis, reasoning, and online learning.
    Figure~\ref{figure:CS:variant_problems_overweighted} provides an illustration of this mechanism.
    
    \item[(ID 13$\sim$15) `Comprehensive']
    This category simultaneously enables the dynamic elements from `Shortsighted Monkey' and `Over-weight Monkey'.
    Specifically, detailed box information becomes unavailable once it leaves the interaction range, and boxes may be crushed with a certain probability when the monkey climbs onto them.
    These scenes are designed to comprehensively evaluate an agent's ability to analyze and respond to uncertainty in highly complex environments. \\
    Moreover, since the probabilistic side effects of crushing boxes can frequently alter box heights, related information stored in the agent's memory may become invalid.
    As a result, this category additionally evaluates the agent's ability to assess the validity of its memory and proactively update it.
\end{description}

\subsection{Comparison Agents}

In this experiment, a total of six different agents are evaluated.
One agent, referred to as \texttt{Basic}, serves as a baseline and is equipped only with a simple observation–action loop.
Two additional agents are reproductions of widely adopted agent architectures, namely \texttt{ReAct} and \texttt{MLDT}.
The remaining three agents are newly proposed in this work: `On-demand Reasoning' (\texttt{OR}), `On-demand Reasoning + Notes' (\texttt{ORN}), and `On-demand Reasoning + Notes + Planning' (\texttt{PlanORN}).

The formal representations of \texttt{ReAct} and \texttt{MLDT} under the Structural Context Model have already been presented in Section~\ref{section:SCM:comparing_agents}.
In the following, we provide the corresponding formal representations for the remaining agents.

Denote the problem description and rules as $\alpha$, action lists as $\beta$.
For the $i$-th step, denote the current status as $\gamma_i$.
Denote the agent memory pattern as $M(\;|\;\mathcal{A})$ from Section~\ref{section:SCM:agent_memory}.

Equation~\ref{equation:CS:Basic} presents the formal expression of \texttt{Basic}.

\begin{equation}
\label{equation:CS:Basic}
\begin{aligned}
    F_{\text{basic}}(\gamma_i) & = \alpha \beta \cdot M(\varepsilon \;|\;\mathcal{A}) \cdot \prod_{i} \gamma_i \cdot (\text{``Perform an action.''})
\end{aligned}
\end{equation}

From the formal expression of \texttt{ReAct} given in Equation~\ref{equation:SCM:ReAct}, we observe that the \texttt{ReAct} agent performs reasoning at every step within a sub-session under the same context, regardless of whether the agent actually has uncertainty about the next action or about events occurring in the current Scene.
This design substantially increases per-step latency and token consumption.
Moreover, because consecutive reasoning processes are performed in close proximity in the context, adjacent reasoning outputs are likely to interfere with each other, thereby reducing the overall information gain.
A natural direction for improvement is therefore to transform reasoning from a mandatory per-step operation into an optional one.

Based on this idea, we propose the `On-demand Reasoning' (\texttt{OR}) agent.
Equation~\ref{equation:CS:OR} presents its formal expression, with the main differences from \texttt{ReAct} highlighted in \textcolor{red}{red}.

Here, the reasoning sub-agent is implemented as an agent tool $A_{\text{reasoning}}$, whose parameter $q_i$ represents the main agent's question about the current context or the instruction that the reasoning process is expected to fulfill.
From this formal expression, we can also directly observe that, in contrast to the \texttt{ReAct} agent, which injects only the reasoning output into the action context, the proposed \texttt{OR} agent additionally injects the question itself into the main agent's action context.

\begin{equation}
\label{equation:CS:OR}
\begin{gathered}
    A_{\text{reasoning}}(\textcolor{red}{q_i},\gamma_i) = \mathbf{S}\bigl(\alpha\beta \cdot \gamma_i \cdot (\text{``... answer the question:''}) \cdot \textcolor{red}{q_i} \bigr)
    \rightarrow \textcolor{red}{r_i} \\
    \begin{aligned}
        F_{\text{OR}}(\gamma_i) & = \alpha \beta \cdot M(\varepsilon \;|\;\mathcal{A}) \cdot \prod_{i} \gamma_i \cdot (\text{``...''}) \cdot \textcolor{red}{\{q_i \cdot A_{\text{reasoning}}(q, \gamma_i)\}^{k_i}} \\
        & = \alpha \beta \cdot M(\varepsilon \;|\;\mathcal{A}) \cdot \prod_{i} \gamma_i \cdot (\text{``...''}) \cdot \textcolor{red}{\{q_i \cdot r_i\}^{k_i}}, \quad k_i \in \{0, 1\}
    \end{aligned}
\end{gathered}
\end{equation}

To enhance memory capability, we propose a variant agent that augments \texttt{OR} with a memory module implemented as a tool, referred to as \texttt{On-demand Reasoning + Notes} (\texttt{ORN}).
The set of notes is denoted as $\mathcal{N} = \{n_1, n_2, ... n_m\}$, Two agent tools, $A_{\text{add-tool}}(n)$ and $A_{\text{remove-tool}}(n)$, are provided to add notes to and remove notes from $\mathcal{N}$, respectively.
According to prior studies on the attention mechanisms of LLMs~\cite{hong2025context_rot, laban2025llms_get_lost_in_converstion}, these models tend to allocate more attention to tokens near the beginning and the end of the context, while information located in the middle of long contexts is more likely to be overlooked or forgotten.
Motivated by this observation, the Notes tool injects the contents of $\mathcal{N}$ into the end of the action context before each action step.
Compared to having such critical information scattered within a long context, this design makes the notes more likely to receive the model's attention.
The formal expression of \texttt{ORN} is given in Equation~\ref{equation:CS:ORN}.
The differences from the \texttt{OR} agent are likewise highlighted in \textcolor{red}{red}.

\begin{equation}
\label{equation:CS:ORN}
\begin{gathered}
    \begin{aligned}
        F_{\text{ORN}}(\gamma_i) & = \alpha \beta \cdot M(\varepsilon \;|\;\mathcal{A}) \cdot \prod_{i} \gamma_i \cdot (\text{``...''}) \cdot \{q_i \cdot A_{\text{reasoning}}(q, \gamma_i)\}^{k_i} \cdot \textcolor{red}{\mathcal{N}} \\
        & = \alpha \beta \cdot M(\varepsilon \;|\;\mathcal{A}) \cdot \prod_{i} \gamma_i \cdot (\text{``...''}) \cdot \{q_i \cdot r_i\}^{k_i} \cdot \textcolor{red}{
        \prod_{n_j \in \mathcal{N}} n_j
        }, \quad k_i \in \{0, 1\}
    \end{aligned}
\end{gathered}
\end{equation}

As analyzed in Section~\ref{section:SCM:comparing_agents}, \texttt{MLDT} produces a static plan.
A natural direction for improvement is therefore to encapsulate the task decomposition and plan generation sub-agent as a tool that can be invoked by the main agent.
This tool is denoted as $A_{plan}(g_i)$, where the parameter $g_i$ represents the agent's feedback on the current plan or the objective for generating a new plan.
Equation~\ref{equation:CS:PlanORN} presents the formal expression of \texttt{PlanORN}, with the differences from \texttt{ORN} highlighted in red.
The prompt used for plan generation is reused from the Semantic Dynamics Analysis results presented in Section~\ref{section:SDA:demonstration_task_decomposition}.

\begin{equation}
\label{equation:CS:PlanORN}
\begin{gathered}
    \begin{aligned}
    \textcolor{red}{A_{plan}(g_i)} = \textcolor{red}{\mathcal{P}_i} = \, & 
    \begin{cases}
        \mathbf{S\bigl(\alpha\beta \cdot \gamma_i \cdot (\text{``Generate a plan...''})\bigr)}, & \text{called in step $i$} \\
        \mathcal{P}_{i - 1}, & \text{not called in step $i$} \\
    \end{cases} 
    \end{aligned} \\
    \begin{aligned}
        F_{\text{PlanORN}}(\gamma_i) = & \alpha \beta \cdot M(\varepsilon \;|\;\mathcal{A}) \cdot
        \prod_{i} \gamma_i \cdot (\text{``...''}) \cdot 
        \textcolor{red}{\textcolor{red}{\mathcal{P}_i}} \cdot 
        \{q_i \cdot A_{\text{reasoning}}(q, \gamma_i)\}^{k_i} \cdot \mathcal{N} \\
        = & \alpha \beta \cdot M(\varepsilon \;|\;\mathcal{A}) \cdot 
        \prod_{i} \gamma_i \cdot (\text{``...''}) \cdot
        \textcolor{red}{\textcolor{red}{\mathcal{P}_i}}
        \cdot \{q_i \cdot r_i\}^{k_i} \cdot
        \prod_{n_j \in \mathcal{N} n_j}, \\
        & k_i \in \{0, 1\}
    \end{aligned}
\end{gathered}
\end{equation}

\subsection{Evaluation Results}

For the 15 scenes and 6 agents, each agent is evaluated on each scene over 100 independent trials.
Each time the agent issues a control command for the monkey (such as horizontal movement, climbing up, climbing down, grabbing a box, or placing a box), it is counted as one step.
A trial is considered successful if the agent controls the monkey to reach the position of the banana within 300 steps.
Otherwise, the trial is regarded as a failure if the step limit of 300 is exceeded or if the agent explicitly invokes the \texttt{Abandon} agent tool.

In these experiments, we collect and analyze following metrics: (1) success rate; (2) three average indicators computed over all trials (including failures): average steps, average time cost, and average token cost; (3) two specialized metrics, \textit{Time-to-Success} and \textit{Token-to-Success}.

Denote the success rate as $r$, denote the average value of a given metrics $X$ (time consumption, token consumption, etc.) over successful trials as $x_s$ , and denote the corresponding average value over failed trials as $x_f$.
The definition of \textit{X-to-Success} is given in Equation~\ref{equation:CS:x_to_success}.

\begin{equation}
\label{equation:CS:x_to_success}
\begin{aligned}
\text{X-to-Success} = x_s + (\frac{1}{r} - 1) \cdot x_f
\end{aligned}
\end{equation}

Conventional average metrics (e.g., success rate and average number of steps) reflect the expected performance of an agent under a fixed number of attempts.
Different metrics correspond to different real-world considerations: for instance, when the cost of failure dominates, success rate is the primary reference, whereas when the cost of each action is high, the average number of steps becomes more relevant.

In contrast, Time-to-Success and Token-to-Success provide an alternative perspective that is more aligned with practical LLM-powered robot agents, where failures are typically inevitable but acceptable (or recoverable), and the user’s instruction only needs to be successfully completed once.
These metrics characterize the expected waiting time and LLM token cost incurred by a user.

Table~\ref{table:CS:experiment_results_classic} presents the experimental results on the \textit{Classic} scene category (Scene IDs ranging from 1 to 3).
Table~\ref{table:CS:experiment_results_dual} presents the experimental results on the \textit{Dual Bananas} scene category (Scene IDs ranging from 4 to 6).
Table~\ref{table:CS:experiment_results_shortsighted} presents the experimental results on the \textit{Shortsighted Monkey} scene category (Scene IDs ranging from 7 to 9).
Table~\ref{table:CS:experiment_results_overweight} presents the experimental results on the \textit{Over-weight Monkey} scene category (Scene IDs ranging from 10 to 12).
Table~\ref{table:CS:experiment_results_comprehensive} presents the experimental results on the \textit{Comprehensive} scene category (Scene IDs ranging from 13 to 15).

\begin{table*}[htbp]
\caption{Experimental results on the `\textit{Classic}' scene category.
The table reports the success rate (SR.$\uparrow$), average number of steps (Avg. Steps$\downarrow$), average execution time (Avg. Time$\downarrow$), average token consumption (Avg. Tokens$\downarrow$), as well as the Time-to-Success (Time-to-S$\downarrow$) and Token-to-Success (Token-to-S$\downarrow$) metrics for each agent.
Agents proposed in this work are marked with an asterisk (*), and the best result for each metric is \underline{underlined}.
Arrows ($\uparrow$/$\downarrow$) indicate whether higher or lower values are better, respectively.
The symbol `$\infty$’ indicates that no successful trials were observed.}
\label{table:CS:experiment_results_classic}
\centering
\adjustbox{max width=0.9\textwidth}{
\begin{tabular}{cccccccc}
\toprule
Scene & Agent & SR.$\uparrow$ & Avg. Steps$\downarrow$ & Avg. Time(s)$\downarrow$ & Avg. Tokens$\downarrow$ & Time-to-S$\downarrow$ & Tokens-to-S$\downarrow$ \\
\midrule
\multirow{6}{*}{Scene 1}
 & Basic    & 0.99 & \ul{7.0} & \ul{14.05} & \ul{11.6K} & \ul{14.19} & \ul{11.8K} \\
 & ReAct    & 0.99 & 7.5 & 40.09 & 17.0K & 40.49 & 17.3K \\
 & MLDT     & 0.98 & 7.3 & 19.85 & 15.0K & 20.25 & 15.6K \\
 & OR*      & \ul{1.00} & 7.1 & 14.88 & 12.7K & 14.88 & 12.7K \\
 & ORN*     & 0.97 & 7.7 & 18.80 & 21.9K & 19.38 & 23.0K \\
 & PlanORN* & \ul{1.00} & 9.0 & 38.97 & 30.7K & 38.97 & 30.7K \\
\midrule

\multirow{6}{*}{Scene 2}
 & Basic    & 0.86 & 23.6 & \ul{45.74} & 212.2K & \ul{53.18} & 121.3K \\
 & ReAct    & 0.87 & 30.9 & 232.64 & 596.9K & 267.40 & 292.7K \\
 & MLDT     & 0.85 & 28.6 & 65.51 & 359.5K & 77.07 & \ul{96.6K} \\
 & OR*      & \ul{0.97} & \ul{22.5} & 61.65 & \ul{205.5K} & 63.56 & 212.9K \\
 & ORN*     & 0.81 & 29.1 & 109.63 & 343.3K & 135.34 & 368.2K \\
 & PlanORN* & 0.85 & 34.1 & 166.60 & 750.2K & 196.00 & 493.3K \\
\midrule

\multirow{6}{*}{Scene 3}
 & Basic    & 0.00 & \ul{27.1} & \ul{49.42} & \ul{186.8K} & $\infty$ & $\infty$ \\
 & ReAct    & 0.53 & 84.3 & 1066.64 & 4.5M & 2012.53 & 910.9K \\
 & MLDT     & 0.00 & 28.1 & 63.27 & 239.0K & $\infty$ & $\infty$ \\
 & OR*      & 0.37 & 70.6 & 291.63 & 2.1M & 788.18 & 4.6M \\
 & ORN*     & 0.39 & 72.9 & 423.08 & 3.0M & 1084.82 & 7.0M \\
 & PlanORN* & \ul{0.66} & 83.8 & 503.38 & 4.2M & \ul{762.69} & \ul{3.2M} \\
\bottomrule
\end{tabular}
}
\end{table*}

\begin{table*}[htbp]
\caption{Experimental results on the `\textit{Dual Bananas}' scene category.
The table reports the success rate (SR.$\uparrow$), average number of steps (Avg. Steps$\downarrow$), average execution time (Avg. Time$\downarrow$), average token consumption (Avg. Tokens$\downarrow$), as well as the Time-to-Success (Time-to-S$\downarrow$) and Token-to-Success (Token-to-S$\downarrow$) metrics for each agent.
Agents proposed in this work are marked with an asterisk (*), and the best result for each metric is \underline{underlined}.
Arrows ($\uparrow$/$\downarrow$) indicate whether higher or lower values are better, respectively.
The symbol `$\infty$’ indicates that no successful trials were observed.}
\label{table:CS:experiment_results_dual}
\centering
\adjustbox{max width=0.9\textwidth}{
\begin{tabular}{cccccccc}
\toprule
Scene & Agent & SR.$\uparrow$ & Avg. Steps$\downarrow$ & Avg. Time(s)$\downarrow$ & Avg. Tokens$\downarrow$ & Time-to-S$\downarrow$ & Tokens-to-S$\downarrow$ \\
\midrule
\multirow{6}{*}{Scene 4}
 & Basic    & 0.97 & 7.4 & \ul{13.54} & \ul{14.7K} & \ul{13.95} & \ul{15.2K} \\
 & ReAct    & 0.99 & 7.7 & 42.31 & 20.7K & 42.73 & 19.5K \\
 & MLDT     & 0.99 & 7.8 & 20.76 & 19.1K & 20.97 & 19.3K \\
 & OR*      & 0.99 & \ul{7.3} & 15.16 & 15.7K & 15.31 & 15.9K \\
 & ORN*     & 0.99 & 8.1 & 18.76 & 25.8K & 18.95 & 26.2K \\
 & PlanORN* & \ul{1.00} & 9.1 & 33.05 & 34.8K & 33.05 & 34.8K \\
\midrule

\multirow{6}{*}{Scene 5}
 & Basic    & 0.50 & 19.6 & \ul{34.89} & 111.1K & 69.77 & 200.9K \\
 & ReAct    & 0.88 & 28.6 & 247.76 & 605.7K & 281.55 & 246.5K \\
 & MLDT     & 0.65 & \ul{17.3} & 39.31 & \ul{93.9K} & \ul{60.48} & \ul{128.3K} \\
 & OR*      & 0.94 & 22.4 & 62.41 & 179.5K & 66.39 & 185.5K \\
 & ORN*     & 0.93 & 23.0 & 84.25 & 253.1K & 90.59 & 254.9K \\
 & PlanORN* & \ul{0.95} & 42.5 & 203.04 & 1.1M & 213.72 & 1.1M \\
\midrule

\multirow{6}{*}{Scene 6}
 & Basic    & 0.90 & 11.2 & \ul{20.10} & 43.6K & \ul{22.34} & \ul{38.5K} \\
 & ReAct    & \ul{1.00} & 12.4 & 79.82 & 70.9K & 79.82 & 70.9K \\
 & MLDT     & 0.90 & \ul{10.6} & 28.91 & \ul{41.7K} & 32.12 & 41.6K \\
 & OR*      & 0.99 & 11.3 & 26.70 & 51.7K & 26.97 & 48.1K \\
 & ORN*     & \ul{1.00} & 11.6 & 32.81 & 61.5K & 32.81 & 61.5K \\
 & PlanORN* & 0.99 & 19.6 & 78.67 & 188.1K & 79.47 & 182.0K \\
\bottomrule
\end{tabular}
}
\end{table*}

\begin{table*}[htbp]
\caption{Experimental results on the `\textit{Shortsighted Monkey}' scene category.
The table reports the success rate (SR.$\uparrow$), average number of steps (Avg. Steps$\downarrow$), average execution time (Avg. Time$\downarrow$), average token consumption (Avg. Tokens$\downarrow$), as well as the Time-to-Success (Time-to-S$\downarrow$) and Token-to-Success (Token-to-S$\downarrow$) metrics for each agent.
Agents proposed in this work are marked with an asterisk (*), and the best result for each metric is \underline{underlined}.
Arrows ($\uparrow$/$\downarrow$) indicate whether higher or lower values are better, respectively.
The symbol `$\infty$’ indicates that no successful trials were observed.}
\label{table:CS:experiment_results_shortsighted}
\centering
\adjustbox{max width=0.9\textwidth}{
\begin{tabular}{cccccccc}
\toprule
Scene & Agent & SR.$\uparrow$ & Avg. Steps$\downarrow$ & Avg. Time(s)$\downarrow$ & Avg. Tokens$\downarrow$ & Time-to-S$\downarrow$ & Tokens-to-S$\downarrow$ \\
\midrule
\multirow{6}{*}{Scene 7}
 & Basic    & \ul{1.00} & \ul{7.6} & \ul{15.05} & \ul{14.1K} & \ul{15.05} & \ul{14.1K} \\
 & ReAct    & 0.99 & 8.7 & 53.46 & 24.2K & 54.00 & 24.4K \\
 & MLDT     & 0.96 & 8.4 & 23.33 & 20.3K & 24.30 & 21.6K \\
 & OR*      & 1.00 & \ul{7.6} & 16.15 & 15.3K & 16.15 & 15.3K \\
 & ORN*     & 0.99 & 8.3 & 20.71 & 25.3K & 20.92 & 25.4K \\
 & PlanORN* & 0.95 & 9.9 & 39.45 & 48.7K & 41.52 & 42.6K \\
\midrule

\multirow{6}{*}{Scene 8}
 & Basic    & 0.66 & 38.0 & \ul{78.81} & 466.9K & 119.41 & 338.8K \\
 & ReAct    & 0.86 & 32.4 & 311.66 & 814.8K & 361.24 & 215.7K \\
 & MLDT     & 0.60 & 36.8 & 82.44 & 532.8K & 137.40 & \ul{182.5K} \\
 & OR*      & 0.94 & 29.6 & 83.31 & \ul{264.3K} & \ul{88.62} & 278.4K \\
 & ORN*     & 0.88 & 26.2 & 96.63 & 279.4K & 109.80 & 293.3K \\
 & PlanORN* & \ul{1.00} & \ul{22.4} & 94.34 & 384.0K & 94.34 & 384.0K \\
\midrule

\multirow{6}{*}{Scene 9}
 & Basic    & 0.00 & 31.6 & 63.09 & 456.4K & $\infty$ & $\infty$ \\
 & ReAct    & 0.50 & 105.9 & 1641.00 & 7.0M & 3251.04 & \ul{1.9M} \\
 & MLDT     & 0.00 & \ul{21.9} & \ul{53.16} & \ul{172.7K} & $\infty$ & $\infty$ \\
 & OR*      & 0.40 & 87.7 & 390.94 & 3.4M & \ul{977.36} & 4.5M \\
 & ORN*     & 0.32 & 133.7 & 1077.15 & 10.4M & 3366.08 & 9.8M \\
 & PlanORN* & \ul{0.56} & 109.7 & 829.00 & 9.0M & 1480.35 & 6.3M \\
\bottomrule
\end{tabular}
}
\end{table*}

\begin{table*}[htbp]
\caption{Experimental results on the `\textit{Over-weight Monkey}' scene category.
The table reports the success rate (SR.$\uparrow$), average number of steps (Avg. Steps$\downarrow$), average execution time (Avg. Time$\downarrow$), average token consumption (Avg. Tokens$\downarrow$), as well as the Time-to-Success (Time-to-S$\downarrow$) and Token-to-Success (Token-to-S$\downarrow$) metrics for each agent.
Agents proposed in this work are marked with an asterisk (*), and the best result for each metric is \underline{underlined}.
Arrows ($\uparrow$/$\downarrow$) indicate whether higher or lower values are better, respectively.
The symbol `$\infty$’ indicates that no successful trials were observed.}
\label{table:CS:experiment_results_overweight}
\centering
\adjustbox{max width=0.9\textwidth}{
\begin{tabular}{cccccccc}
\toprule
Scene & Agent & SR.$\uparrow$ & Avg. Steps$\downarrow$ & Avg. Time(s)$\downarrow$ & Avg. Tokens$\downarrow$ & Time-to-S$\downarrow$ & Tokens-to-S$\downarrow$ \\
\midrule
\multirow{6}{*}{Scene 10}
 & Basic    & 0.94 & 11.9 & \ul{22.12} & 37.3K & \ul{23.53} & 38.8K \\
 & ReAct    & \ul{1.00} & 10.8 & 80.51 & 45.3K & 80.51 & 45.3K \\
 & MLDT     & 0.94 & \ul{9.7} & 25.78 & \ul{30.8K} & 27.43 & \ul{33.4K} \\
 & OR*      & 0.98 & 11.4 & 31.03 & 41.8K & 31.66 & 43.5K \\
 & ORN*     & 0.90 & 17.3 & 53.95 & 119.9K & 59.95 & 120.2K \\
 & PlanORN* & 0.96 & 15.1 & 62.13 & 121.7K & 64.72 & 122.3K \\
\midrule

\multirow{6}{*}{Scene 11}
 & Basic    & 0.30 & 22.2 & \ul{41.97} & 117.7K & 139.90 & 322.9K \\
 & ReAct    & 0.57 & 67.5 & 781.70 & 2.3M & 1371.41 & 1.3M \\
 & MLDT     & 0.40 & \ul{20.5} & 45.38 & \ul{111.7K} & \ul{113.46} & \ul{277.6K} \\
 & OR*      & 0.82 & 35.4 & 109.43 & 437.9K & 133.45 & 376.1K \\
 & ORN*     & 0.47 & 40.7 & 172.64 & 739.4K & 367.32 & 1.4M \\
 & PlanORN* & \ul{0.67} & 52.8 & 324.53 & 2.2M & 484.38 & 2.7M \\
\midrule

\multirow{6}{*}{Scene 12}
 & Basic    & 0.00 & \ul{25.7} & \ul{46.49} & \ul{189.8K} & $\infty$ & $\infty$ \\
 & ReAct    & 0.33 & 106.9 & 1541.90 & 5.5M & 4671.04 & \ul{7.0M} \\
 & MLDT     & 0.00 & 26.6 & 62.15 & 213.3K & $\infty$ & $\infty$ \\
 & OR*      & 0.26 & 78.5 & 385.49 & 3.0M & 1482.66 & 8.3M \\
 & ORN*     & 0.22 & 66.3 & 390.96 & 2.9M & 1777.07 & 21.4M \\
 & PlanORN* & \ul{0.50} & 99.6 & 726.11 & 6.8M & \ul{1452.21} & 10.2M \\
\bottomrule
\end{tabular}
}
\end{table*}

\begin{table*}[htbp]
\caption{Experimental results on the `\textit{Comprehensive}' scene category.
The table reports the success rate (SR.$\uparrow$), average number of steps (Avg. Steps$\downarrow$), average execution time (Avg. Time$\downarrow$), average token consumption (Avg. Tokens$\downarrow$), as well as the Time-to-Success (Time-to-S$\downarrow$) and Token-to-Success (Token-to-S$\downarrow$) metrics for each agent.
Agents proposed in this work are marked with an asterisk (*), and the best result for each metric is \underline{underlined}.
Arrows ($\uparrow$/$\downarrow$) indicate whether higher or lower values are better, respectively.
The symbol `$\infty$’ indicates that no successful trials were observed.}
\label{table:CS:experiment_results_comprehensive}
\centering
\adjustbox{max width=0.9\textwidth}{
\begin{tabular}{cccccccc}
\toprule
Scene & Agent & SR.$\uparrow$ & Avg. Steps$\downarrow$ & Avg. Time(s)$\downarrow$ & Avg. Tokens$\downarrow$ & Time-to-S$\downarrow$ & Tokens-to-S$\downarrow$ \\
\midrule
\multirow{6}{*}{Scene 13}
 & Basic    & 0.97 & 8.5 & \ul{16.48} & \ul{19.3K} & \ul{16.99} & \ul{19.2K} \\
 & ReAct    & 0.99 & 12.9 & 109.50 & 173.0K & 110.60 & 57.0K \\
 & MLDT     & 0.98 & \ul{8.3} & 22.54 & 20.8K & 23.00 & 21.6K \\
 & OR*      & \ul{1.00} & 8.7 & 20.03 & 21.9K & 20.03 & 21.9K \\
 & ORN*     & 0.93 & 13.5 & 40.30 & 70.5K & 43.34 & 72.2K \\
 & PlanORN* & 0.94 & 13.2 & 51.40 & 91.7K & 54.68 & 92.7K \\
\midrule

\multirow{6}{*}{Scene 14}
 & Basic    & 0.03 & 21.6 & 36.91 & 132.4K & 1230.25 & 7.0M \\
 & ReAct    & 0.68 & 71.8 & 856.34 & 2.9M & 1259.33 & 1.5M \\
 & MLDT     & 0.12 & \ul{18.4} & 43.40 & \ul{110.2K} & 361.66 & 1.2M \\
 & OR*      & 0.78 & 40.7 & 133.28 & 579.1K & \ul{170.87} & \ul{628.3K} \\
 & ORN*     & 0.72 & 54.5 & 276.57 & 1.7M & 384.13 & 1.7M \\
 & PlanORN* & \ul{0.89} & 48.0 & 253.78 & 1.6M & 285.15 & 1.3M \\
\midrule

\multirow{6}{*}{Scene 15}
 & Basic    & 0.00 & \ul{21.9} & \ul{41.24} & \ul{174.8K} & $\infty$ & $\infty$ \\
 & ReAct    & 0.18 & 120.1 & 1880.30 & 7.4M & 10446.09 & \ul{16.3M} \\
 & MLDT     & 0.00 & 23.3 & 60.95 & 235.8K & $\infty$ & $\infty$ \\
 & OR*      & 0.27 & 111.9 & 691.11 & 6.4M & 2585.26 & 17.4M \\
 & ORN*     & 0.24 & 129.0 & 1077.71 & 10.9M & 4490.45 & 30.4M \\
 & PlanORN* & \ul{0.32} & 115.6 & 793.93 & 8.5M & \ul{2481.03} & 17.9M \\
\bottomrule
\end{tabular}
}
\end{table*}

\begin{figure}[htbp]
\centering
\includegraphics[width=0.9\textwidth]{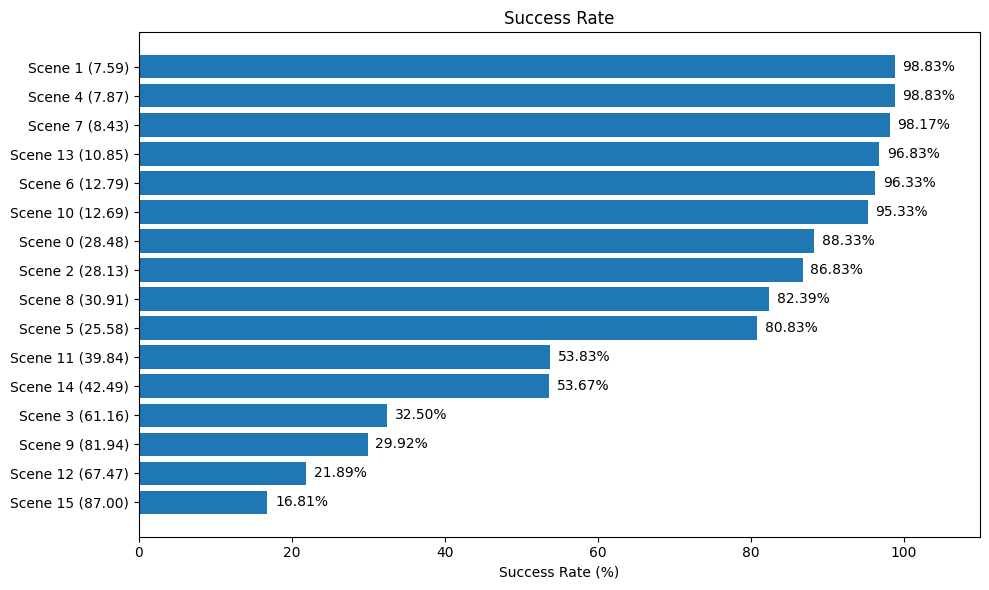}
\caption{
Ranking of scene difficulty based on success rates across all agents.
Scenes are ordered in ascending difficulty from top to bottom.
When multiple scenes share the same average success rate, they are further ordered by the average number of steps in ascending order.
The horizontal axis indicates the success rate, while the average number of steps for each scene is shown in parentheses next to the corresponding scene ID.
}
\label{figure:CS:scene_difficulties}
\end{figure}

We estimate the difficulty of each scene by jointly considering the success rates and average numbers of steps achieved by all agents across scenes.
Based on this estimation, the scenes are ranked in ascending order of difficulty, as shown in Figure~\ref{figure:CS:scene_difficulties}.
The horizontal axis corresponds to the scene IDs, with the average number of steps indicated in parentheses.
The performance of these agents on the five most difficult scenes (ID 14, ID 3, ID 9, ID 12, ID 15) is summarized in Table~\ref{table:appendix:experiment_results_most_difficult}.

From this table, we observe that the 15 scenes exhibit clear separability in terms of difficulty.
They can be roughly grouped into four difficulty intervals, corresponding to success-rate ranges of 98.83\%$\sim$95.33\%, 88.33\%$\sim$80.83\%, 53.83\%$\sim$53.67\%, and 32.50\%$\sim$16.81\%, respectively.
Within each interval, the difficulty increases relatively smoothly, enabling a comprehensive evaluation of performance differences among agents.
In contrast, the final interval shows a much steeper increase in difficulty, thereby more strongly distinguishing agents that exhibit greater robustness under abrupt increases in task difficulty.

\begin{figure}[htbp]
\centering
\includegraphics[width=0.9\textwidth]{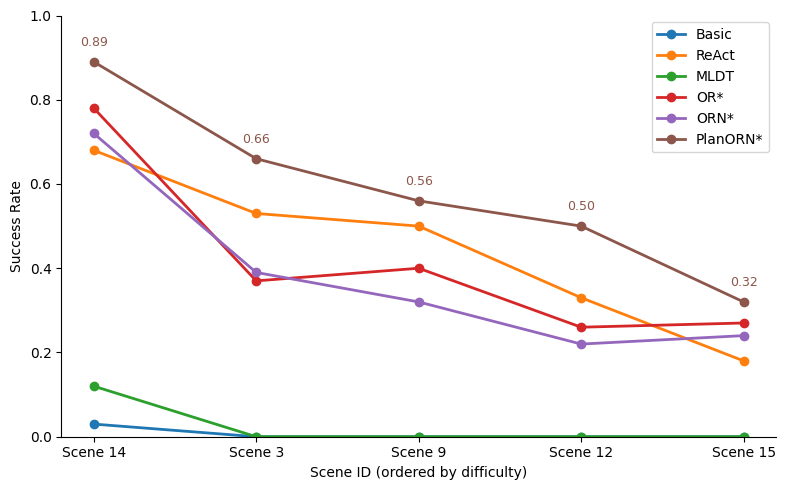}
\caption{
Success rates of different agents on the five most difficult scenes, ordered by increasing difficulty.
}
\label{figure:CS:success_rates_in_hardest_scenes}
\end{figure}

Figure~\ref{figure:CS:success_rates_in_hardest_scenes} compares the success rates of all agents on the five most difficult scenes, ordered by increasing difficulty.
As the scene difficulty increases, a consistent performance degradation can be observed across all agents.

As shown in the figure, both the \texttt{ReAct} agent and the proposed \texttt{PlanORN} agent exhibit relatively strong resilience as scene difficulty increases.
Notably, the success rate of \texttt{PlanORN} consistently exceeds that of \texttt{ReAct} across all five scenes.
In addition, although the \texttt{OR} and \texttt{ORN} agents experience a pronounced drop in success rate from Scene~14 to Scene~3, their performance subsequently stabilizes at a relatively consistent suboptimal level, indicating a certain degree of robustness under further increases in difficulty.

\subsection{Result Analysis}

In this subsection, we summarize the performance of each agent on the proposed benchmark and analyze the underlying reasons for their observed behaviors from the perspective of context structure.

Overall, \texttt{PlanORN} achieves the strongest performance across the entire benchmark, exhibiting clear advantages in success rates on medium- and high-difficulty scenes and demonstrating strong resilience as task difficulty increases.
In contrast, while its capability is limited to simpler settings, \texttt{Basic} offers overwhelming cost efficiency on easy scenes, making it a highly competitive baseline when task complexity is low.

\subsubsection{`\texttt{Basic}' Agent}
\texttt{Basic} is the only agent among the six that does not incorporate any sub-agent component, i.e., its formal expression does not contain $\mathbf{S}(...)$.
As a purely reactive baseline, it exhibits extremely low time and token consumption in simple scenes where the environment is fully observable and deterministic.
However, once the scene slightly violates these assumptions, its success rate drops sharply, indicating a lack of robustness to uncertainty and delayed effects.

\subsubsection{\texttt{ReAct} Agent}
\texttt{ReAct} augments the \texttt{Basic} agent with an explicit reasoning sub-agent, enabling iterative reflection and action revision.
This mechanism improves robustness in moderately difficult scenes but incurs substantial computational overhead, resulting in significantly increased time and token consumption as scene complexity grows.

\subsubsection{\texttt{MLDT} Agent}
\texttt{MLDT} exhibits a largely mediocre overall performance across the evaluated scenes. Although its task decomposition sub-agent produces an explicit static plan, this plan does not translate into a success-rate advantage in difficult scenarios. In particular, when classical planning assumptions are violated, the success rate of \texttt{MLDT} drops to nearly zero, indicating that the static plan is insufficient to handle uncertainty, partial observability, or irreversible side effects.

Nevertheless, the same static planning mechanism enables \texttt{MLDT} to terminate unsuccessful attempts more quickly once the precomputed plan becomes infeasible. As a result, in several moderately difficult scenes (e.g., Scenes 5, 8, 10, and 11), \texttt{MLDT} achieves lower average execution time and, in some cases, smaller Tokens-to-Success compared to more adaptive agents. This efficiency, however, stems primarily from early abandonment rather than improved problem-solving capability.

Overall, \texttt{MLDT} demonstrates that task decomposition alone, when coupled with a static world assumption, offers limited robustness benefits and cannot support sustained performance in challenging environments.

\subsubsection{\texttt{OR} \& \texttt{ORN} Agents}

\texttt{OR} and \texttt{ORN} represent a middle tier of agent designs that introduce structured online reasoning, with \texttt{ORN} further augmenting the architecture with an explicit note-based memory sub-agent. 
Compared to \texttt{Basic}, \texttt{ReAct}, and \texttt{MLDT}, both agents demonstrate markedly improved robustness across medium and hard scenes, maintaining stable non-zero success rates while incurring substantially lower time and token costs than reasoning-heavy approaches. 
This indicates that structured reasoning alone already provides a meaningful robustness gain under partial observability and delayed effects.

The additional memory component (`notes') in \texttt{ORN} yields only limited improvements over \texttt{OR}, and primarily in a subset of medium-difficulty scenes. 
A key limitation lies in the fact that stored notes may become outdated or invalid as the environment evolves, yet are not always updated or cleared in a timely manner. 
Such stale notes can mislead subsequent reasoning and, in harder scenes, may even degrade performance relative to \texttt{OR}. 
Moreover, advanced LLMs exhibit a certain degree of robustness to `context rot', which further dilutes the marginal benefit brought by explicit notes. 
As a result, \texttt{ORN} improves upon OR only under specific conditions, while failing to provide a consistent advantage in the most challenging scenarios.

\subsubsection{\texttt{PlanORN} Agent}

\texttt{PlanORN} represents the most comprehensive agent design evaluated in this study, combining explicit planning with note-based memory under a unified control structure.
Although its time and token consumption are higher than those of the Basic agent in simple scenes, this overhead does not translate into a disadvantage as scene difficulty increases.
On the contrary, \texttt{PlanORN} exhibits substantially higher stability and success rates in medium- and high-difficulty scenarios, where other agents experience rapid performance degradation.

In particular, as scene complexity increases, the performance of \texttt{PlanORN} degrades in a gradual and controlled manner rather than collapsing abruptly.
This behavior indicates that explicit planning enables the agent to systematically reassess action sequences and memory validity when confronted with partial observability, delayed effects, or irreversible state changes.
As a result, despite incurring additional planning overhead, \texttt{PlanORN} achieves more favorable Time-to-Success and Tokens-to-Success in challenging scenes, making it the most robust and well-balanced agent under the proposed benchmark.


\section{Conclusion}

Research on LLM-based agents has progressed rapidly, yet agent engineering remains challenging in practice, and agent designs are often described in highly heterogeneous and ad-hoc ways.
This diversity of formulations reflects the complexity of agent systems, but it also introduces fundamental difficulties for both research and engineering.
From an academic perspective, the growing engineering complexity of agents tends to obscure discussions of design principles and methodologies, allowing implementation details to leak into and confound higher-level analyses of agent behavior.
From an engineering perspective, the design and improvement of agents are still largely heuristic, relying heavily on developers' experience and intuition rather than systematic and interpretable guidance.

Motivated by these challenges, the goal of this work is to identify an implementation-independent modeling perspective that captures the essential structure of LLM-based agents, and to build upon this perspective a sustainable and interpretable workflow for agent analysis and improvement.
By decoupling agent design from specific implementations, we aim to support clearer methodological reasoning in research and more principled iteration in practice.

To address these challenges, we propose the \texttt{Structural Context Model} as a principled representation for describing LLM-based agents.
In contrast to the ad-hoc and task-specific formulations commonly used in existing work, the model provides a clear boundary on what aspects of an agent are explicitly represented and what aspects are intentionally abstracted away.
This explicit delineation allows agent designs to be discussed at the level of structure, rather than being entangled with implementation details or engineering artifacts.

By modeling agents as precise context structures and their transformations, the \texttt{Structural Context Model} offers a more direct and interpretable view of an agent’s core mechanisms.
The resulting representations are not only concise, but also amenable to formal analysis and comparison, enabling differences between agents to be examined in a systematic and implementation-independent manner.
As a result, agent designs that may appear disparate under surface-level descriptions can be understood and contrasted within a unified analytical framework.

Under the \texttt{Structural Context Model}, agent behavior is expressed in terms of composable functions, referred to as context patterns.
This functional view naturally supports composition and reuse, providing a principled basis for modular agent design.
By abstracting agent components as reusable patterns rather than monolithic implementations, the model facilitates higher reusability of agent components, thereby improving development efficiency and accelerating design iteration.

Building upon this perspective, we further propose \texttt{Semantic Dynamics Analysis} as a methodology for extracting reusable context patterns from existing agent designs and for analyzing their roles and interactions.
Rather than stopping at the decomposition and comparison of agent designs, this methodology leverages the descriptive and analytical capabilities of the \texttt{Structural Context Model} to study how context patterns contribute to agent behavior and how they relate to one another.
Importantly, \texttt{Semantic Dynamics Analysis} serves as a bridge between formal analysis and practical engineering: it enables insights derived from structural analysis to feed back into agent implementation through faster iteration, while also supporting a shift in agent research from heuristic, experience-driven study toward more data-driven and sustainable investigation.

We evaluate the proposed model and methodology in a set of variant monkey–banana problems and use them to guide the development of several agent designs.
The benchmark provides a controlled yet challenging environment for studying agent behavior under increasing task difficulty.
Through this process, we find that the proposed approach substantially improves development and iteration efficiency, enabling the rapid construction of agents that exhibit stronger robustness to difficulty scaling.
In particular, the resulting agents maintain competitive overall engineering efficiency in medium and difficult scenarios, demonstrating the practical effectiveness of the proposed framework.

\subsection{Limitation \& Future Work}

While the \texttt{Structural Context Model} and \texttt{Semantic Dynamics Analysis} provide a principled foundation for analyzing and engineering LLM-based agents, several limitations remain and point to promising directions for future work.

\begin{enumerate}
    \item The current formulation of \texttt{Semantic Dynamics Analysis} offers only a preliminary characterization of the geometric relationships among context patterns.
    Although initial propositions demonstrate the feasibility of analyzing and comparing patterns within the proposed framework, these results remain at an early stage.
    Stronger and more expressive propositions and lemmas are needed to more precisely characterize pattern properties and interactions.
    Developing such results will likely require analysis across a substantially larger and more diverse set of agent designs, enabling the refinement of definitions and the derivation of more general theoretical statements.
    \item The degree of automation supported by the current \texttt{Semantic Dynamics Analysis} remains limited.
    At present, the identification and interpretation of context patterns still rely on significant human involvement.
    If stronger, more formally grounded propositions can be established to automatically detect and classify special properties of context patterns, it may become possible to deploy automated analysis systems at scale.
    Such systems could, for example, perform large-scale mining of agent implementations on public code repositories to discover recurring or structurally significant context patterns.
    \item An open and speculative direction concerns the discovery of novel context patterns that are not yet widely used in existing agent designs.
    Beyond analyzing known patterns, \texttt{Semantic Dynamics Analysis} may offer a pathway to systematically uncover new patterns that are more efficient or effective with respect to specific objectives.
    Whether such patterns can be reliably discovered through data-driven structural analysis remains an open question and an intriguing avenue for future investigation.
\end{enumerate}

Taken together, these limitations highlight that the \texttt{Structural Context Model} and \texttt{Semantic Dynamics Analysis} currently remain at the level of formal representations and initial analytical tools.
A natural next step is to further refine their definitions and introduce stricter and more expressive propositions, with the goal of developing a more complete algebraic system for reasoning about agent designs.
Progress along this direction may help advance the study of LLM-based agents from experience-driven Agent Engineering toward a more data- and proof-driven Agent Science.


\begin{appendices}

\section{Detailed Analysis Results for the Task Decomposition Prompt}
\label{appendix:task_decomposition_result}

\begin{sidewaysfigure}[htb]
    \includegraphics[width=1.0\linewidth]{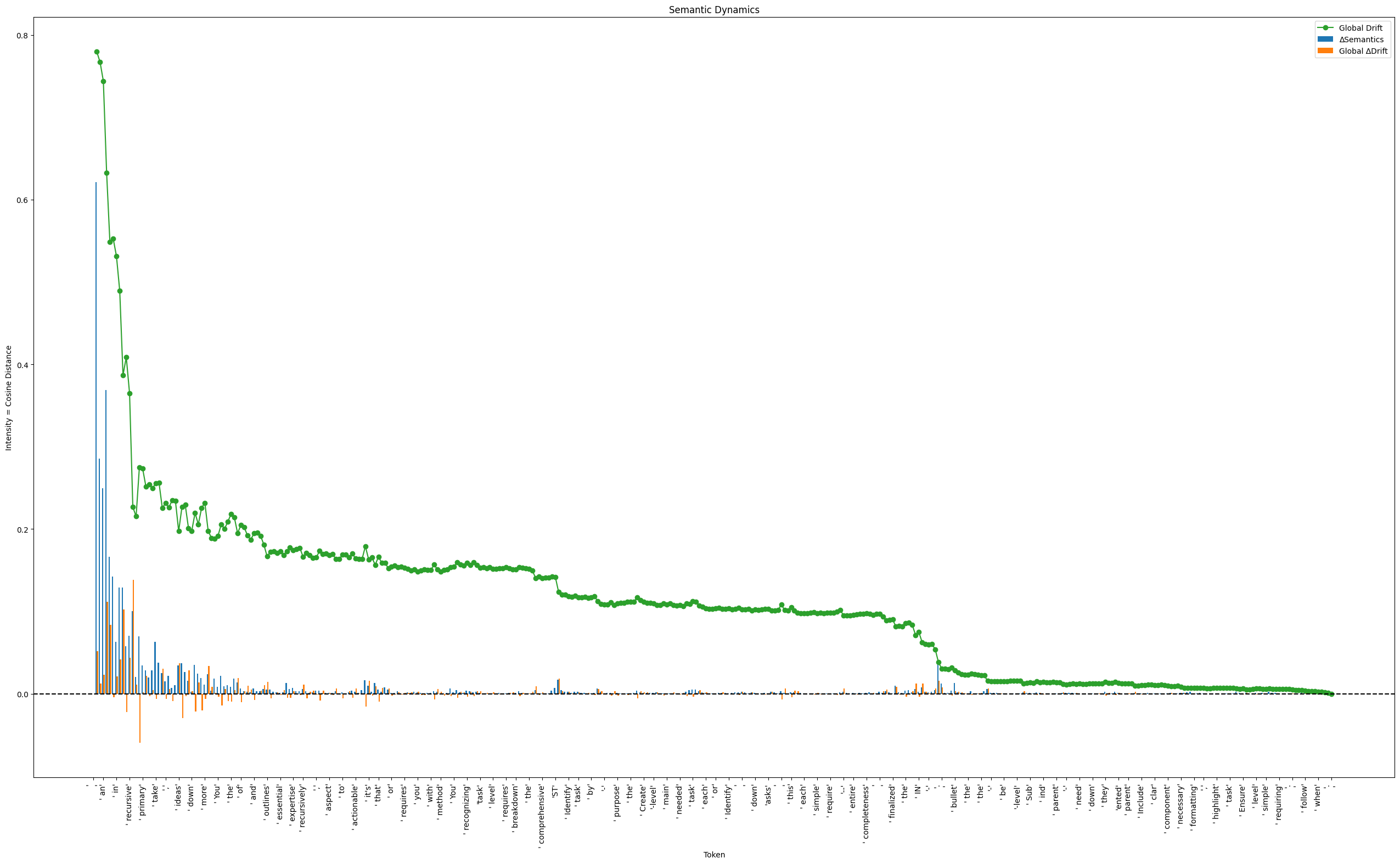}
    \caption{Complete semantic dynamics analysis of a task-decomposition prompt excerpted from the open-source prompt library \textit{Fabric}.}
    \label{figure:appendix:task_decomposition_full_result}
\end{sidewaysfigure}

Figure~\ref{figure:appendix:task_decomposition_full_result} presents the complete semantic dynamics analysis results for the task-decomposition prompt discussed in the main text.

Table~\ref{table:appendix:pairwise_segments_delta_semantics} reports the pairwise $\Delta$Semantics between segments under different ordering configurations, which is used to assess the degree of order invariance between segment pairs.
From the results, the segment pairs (Segment~1, Segment~2), corresponding to the role settings and the goal settings, as well as (Segment~3, Segment~4), corresponding to the detailed steps and the output format, exhibit relatively strong order invariance.

\begin{table}[t]
\caption{Pairwise $\Delta$Semantics Between Segments Under Different Orders}
\label{table:appendix:pairwise_segments_delta_semantics}
\centering
\begin{tabular}{cccc}
    \toprule
    ID & Segment 2 & Segment 3 & Segment 4  \\ 
    \midrule
    Segment1 & 0.3162 & 0.4347 & 0.5283  \\
    Segment2 & - & 0.4584 & 0.5536 \\
    Segment3 & - & - & 0.3776 \\
    \bottomrule
\end{tabular}
\end{table}

\section{Experiments Results on the Most Difficult Scenes}
\label{section:appendix:experiment_results_most_difficult}

Table~\ref{table:appendix:experiment_results_most_difficult} presents the experimental results on the five most difficult scenes (ID 14, ID 3, ID 9, ID 12, ID 15).

\begin{table*}[htbp]
\caption{Experimental results on the five most difficult scenes (ID 14, ID 3, ID 9, ID 12, ID 15).
The table reports the success rate (SR.$\uparrow$), average number of steps (Avg. Steps$\downarrow$), average execution time (Avg. Time$\downarrow$), average token consumption (Avg. Tokens$\downarrow$), as well as the Time-to-Success (Time-to-S$\downarrow$) and Token-to-Success (Token-to-S$\downarrow$) metrics for each agent.
Agents proposed in this work are marked with an asterisk (*), and the best result for each metric is \underline{underlined}.
Arrows ($\uparrow$/$\downarrow$) indicate whether higher or lower values are better, respectively.
The symbol `$\infty$’ indicates that no successful trials were observed.}
\label{table:appendix:experiment_results_most_difficult}
\centering
\adjustbox{max width=0.9\textwidth}{
\begin{tabular}{cccccccc}
\toprule
Scene & Agent & SR.$\uparrow$ & Avg. Steps$\downarrow$ & Avg. Time(s)$\downarrow$ & Avg. Tokens$\downarrow$ & Time-to-S$\downarrow$ & Tokens-to-S$\downarrow$ \\
\midrule
\multirow{6}{*}{Scene 14}
 & Basic    & 0.03 & 21.6 & \ul{36.91} & 132.4K & 1230.25 & 7.0M \\
 & ReAct    & 0.68 & 71.8 & 856.34 & 2.9M & 1259.33 & 1.5M \\
 & MLDT     & 0.12 & \ul{18.4} & 43.40 & \ul{110.2K} & 361.66 & 1.2M \\
 & OR*      & 0.78 & 40.7 & 133.28 & 579.1K & \ul{170.87} & \ul{628.3K} \\
 & ORN*     & 0.72 & 54.5 & 276.57 & 1.7M & 384.13 & 1.7M \\
 & PlanORN* & \ul{0.89} & 48.0 & 253.78 & 1.6M & 285.15 & 1.3M \\
\midrule

\multirow{6}{*}{Scene 3}
 & Basic    & 0.00 & \ul{27.1} & \ul{49.42} & \ul{186.8K} & $\infty$ & $\infty$ \\
 & ReAct    & 0.53 & 84.3 & 1066.64 & 4.5M & 2012.53 & \ul{910.9K} \\
 & MLDT     & 0.00 & 28.1 & 63.27 & 239.0K & $\infty$ & $\infty$ \\
 & OR*      & 0.37 & 70.6 & 291.63 & 2.1M & 788.18 & 4.6M \\
 & ORN*     & 0.39 & 72.9 & 423.08 & 3.0M & 1084.82 & 7.0M \\
 & PlanORN* & \ul{0.66} & 83.8 & 503.38 & 4.2M & \ul{762.69} & 3.2M \\
\midrule

\multirow{6}{*}{Scene 9}
 & Basic    & 0.00 & 31.6 & 63.09 & 456.4K & $\infty$ & $\infty$ \\
 & ReAct    & 0.50 & 105.9 & 1641.00 & 7.0M & 3251.04 & \ul{1.9M} \\
 & MLDT     & 0.00 & \ul{21.9} & \ul{53.16} & \ul{172.7K} & $\infty$ & $\infty$ \\
 & OR*      & 0.40 & 87.7 & 390.94 & 3.4M & \ul{977.36} & 4.5M \\
 & ORN*     & 0.32 & 133.7 & 1077.15 & 10.4M & 3366.08 & 9.8M \\
 & PlanORN* & \ul{0.56} & 109.7 & 829.00 & 9.0M & 1480.35 & 6.3M \\
\midrule

\multirow{6}{*}{Scene 12}
 & Basic    & 0.00 & 25.7 & \ul{46.49} & \ul{189.8K} & $\infty$ & $\infty$ \\
 & ReAct    & 0.33 & 106.9 & 1541.90 & 5.5M & 4671.04 & \ul{7.0M} \\
 & MLDT     & 0.00 & \ul{26.6} & 62.15 & 213.3K & $\infty$ & $\infty$ \\
 & OR*      & 0.26 & 78.5 & 385.49 & 3.0M & 1482.66 & 8.3M \\
 & ORN*     & 0.22 & 66.3 & 390.96 & 2.9M & 1777.07 & 21.4M \\
 & PlanORN* & \ul{0.50} & 99.6 & 726.11 & 6.8M & \ul{1452.21} & 10.2M \\
\midrule

\multirow{6}{*}{Scene 15}
 & Basic    & 0.00 & \ul{21.9} & \ul{41.24} & \ul{174.8K} & $\infty$ & $\infty$ \\
 & ReAct    & 0.18 & 120.1 & 1880.30 & 7.4M & 10446.09 & \ul{16.3M} \\
 & MLDT     & 0.00 & 23.3 & 60.95 & 235.8K & $\infty$ & $\infty$ \\
 & OR*      & 0.27 & 111.9 & 691.11 & 6.4M & 2585.26 & 17.4M \\
 & ORN*     & 0.24 & 129.0 & 1077.71 & 10.9M & 4490.45 & 30.4M \\
 & PlanORN* & \ul{0.32} & 115.6 & 793.93 & 8.5M & \ul{2481.03} & 17.9M \\
\bottomrule
\end{tabular}
}
\end{table*}

\end{appendices}

\bibliography{main}

\end{document}